\documentclass[runningheads]{llncs}
\usepackage{graphicx}
\usepackage{amsmath,amssymb} % define this before the line numbering.
\usepackage{color}
\usepackage{epsfig}
\usepackage{graphicx}
\usepackage{subfig}
\usepackage{soul}
\usepackage{courier}
\usepackage[bookmarks=false]{hyperref}
\usepackage[space]{cite}

\graphicspath{ {images_ds/JPEG/} }

\usepackage[width=122mm,left=12mm,paperwidth=146mm,height=193mm,top=12mm,paperheight=217mm]{geometry}

\usepackage{import}
\newcommand{\specialcell}[2][c]{%
  \begin{tabular}[#1]{@{}c@{}}#2\end{tabular}}
 
\begin{document}
% \linenumbers
\pagestyle{headings}
\mainmatter
\def\ECCV16SubNumber{497}  % Insert your submission number here

%\title{LCrowdV: Generating Labeled Crowd Videos using Procedural Simulation} % Replace with your title
\title{LCrowdV: Generating Labeled Videos for Simulation-based Crowd Behavior Learning} % Replace with your title
 \titlerunning{Generating Labeled Videos for Simulation-based Crowd Behavior Learning}

% \authorrunning{ECCV-16 submission ID \ECCV16SubNumber}

%  \author{Anonymous ECCV submission}
%  \institute{Paper ID \ECCV16SubNumber}

\newcommand*\samethanks[1][\value{footnote}]{\footnotemark[#1]}
%fix footnote numbers
    \makeatletter
    \let\@fnsymbol\@arabic
    \makeatother

\authorrunning{Ernest Cheung, T.K. Wong, Aniket Bera, Xiaogang Wang, Dinesh Manocha}

  \author{Ernest Cheung\thanks {\{ernestc, ⁠⁠⁠ansowong, ab, dm\}@cs.unc.edu} \and Tsan Kwong Wong\samethanks \and Aniket Bera\samethanks \and Xiaogang Wang\thanks{xgwang@ee.cuhk.edu.hk} \and  Dinesh Manocha\samethanks[1]}

 % \url{http://LCrowdV.net}
  \institute{The University of North Carolina at Chapel Hill}

\maketitle

\begin{abstract}
We present a novel procedural framework to generate an arbitrary number of labeled crowd videos (LCrowdV). The resulting crowd video datasets are used to design accurate algorithms or training models for crowded scene understanding. Our overall approach is composed of two components: a procedural simulation framework for generating crowd movements and behaviors, and a procedural rendering framework to generate different videos or images. Each video or image is automatically labeled based on the environment, number of pedestrians, density, behavior, flow, lighting conditions, viewpoint, noise, etc. Furthermore, we can increase the realism by combining synthetically-generated behaviors with real-world background videos. We demonstrate the benefits of LCrowdV over prior lableled crowd datasets by improving the accuracy of pedestrian detection  and crowd behavior classification algorithms.  
LCrowdV would be released on the WWW.

\end{abstract}

\section{Introduction}

The accessibility of commodity cameras has lead to wide availability of crowd videos.  In particular, videos of crowds consisting of tens or hundreds (or more) of human agents or pedestrians are increasingly becoming available on the internet, e.g. YouTube. 
%These videos may correspond to political, religious or sporting events, mobs,  or to movements of pedestrians in a building or on urban streets. 
One of the main challenges in computer vision and related areas is crowded scene understanding or crowd video analysis. These include many sub-problems corresponding to crowd behavior analysis, crowd tracking, crowd segmentation, crowd counting, abnormal behavior detection, crowd  prediction, etc.

The problems related to crowded scene understanding have been extensively studied. Many solutions for each of these sub-problems have been developed by using crowd video datasets~\cite{mikel_dataDriven,rabaud2006counting,hattori2015learning,5597451,shao2015deeply} along with different techniques for computing robust features or learning the models. However, most of these datasets are limited, either in terms of different crowd behavior or scenarios, or the accuracy of the labels. 
%The limitations of current datasets affects the accuracy or reliability of  current solutions for different sub-problems.  
 
\begin{figure}

\vspace{-0.2cm}
\centering
\subfloat[]{\includegraphics[width=0.33\textwidth]{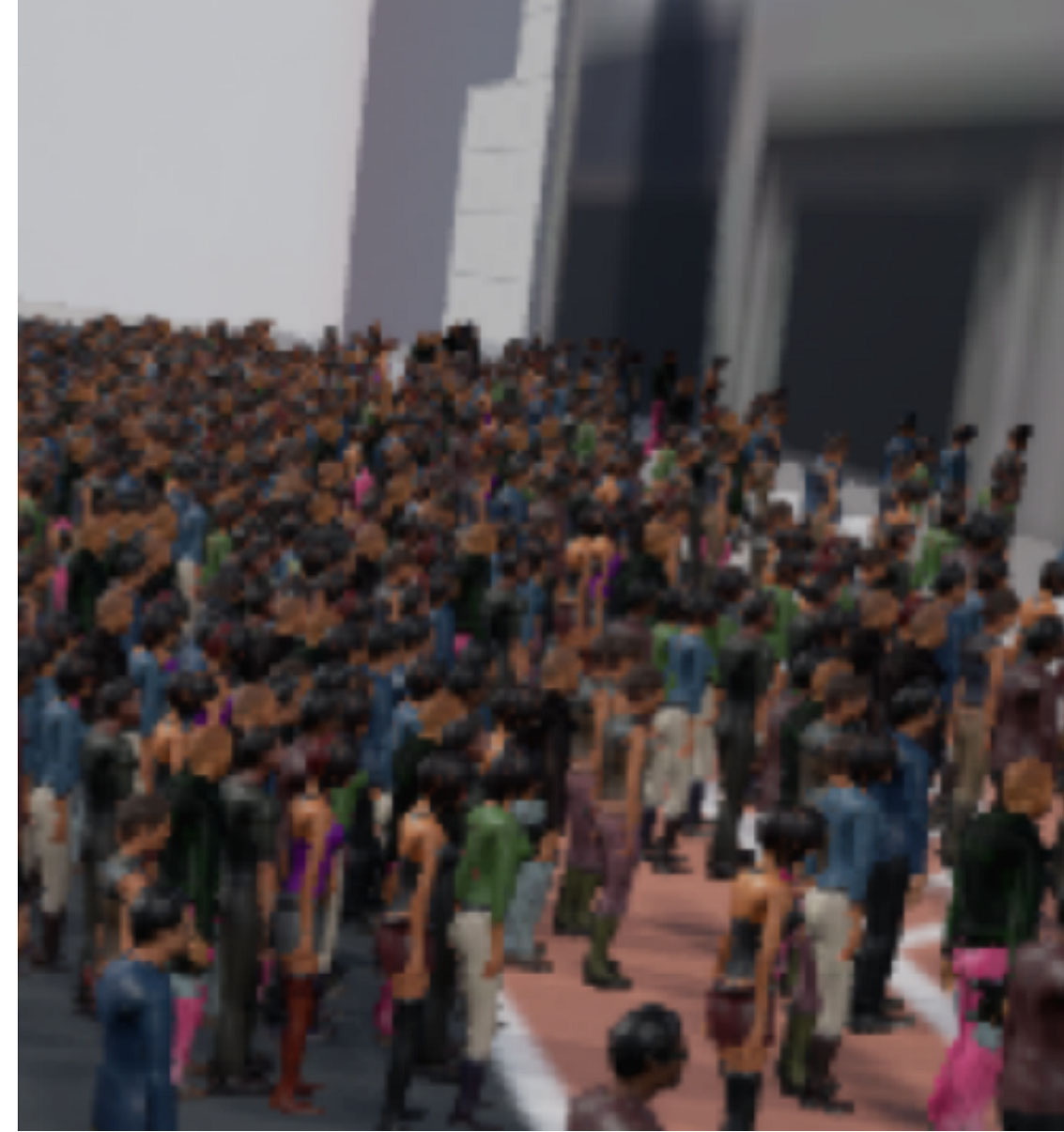}}~
\subfloat[]{\includegraphics[width=0.33\textwidth]{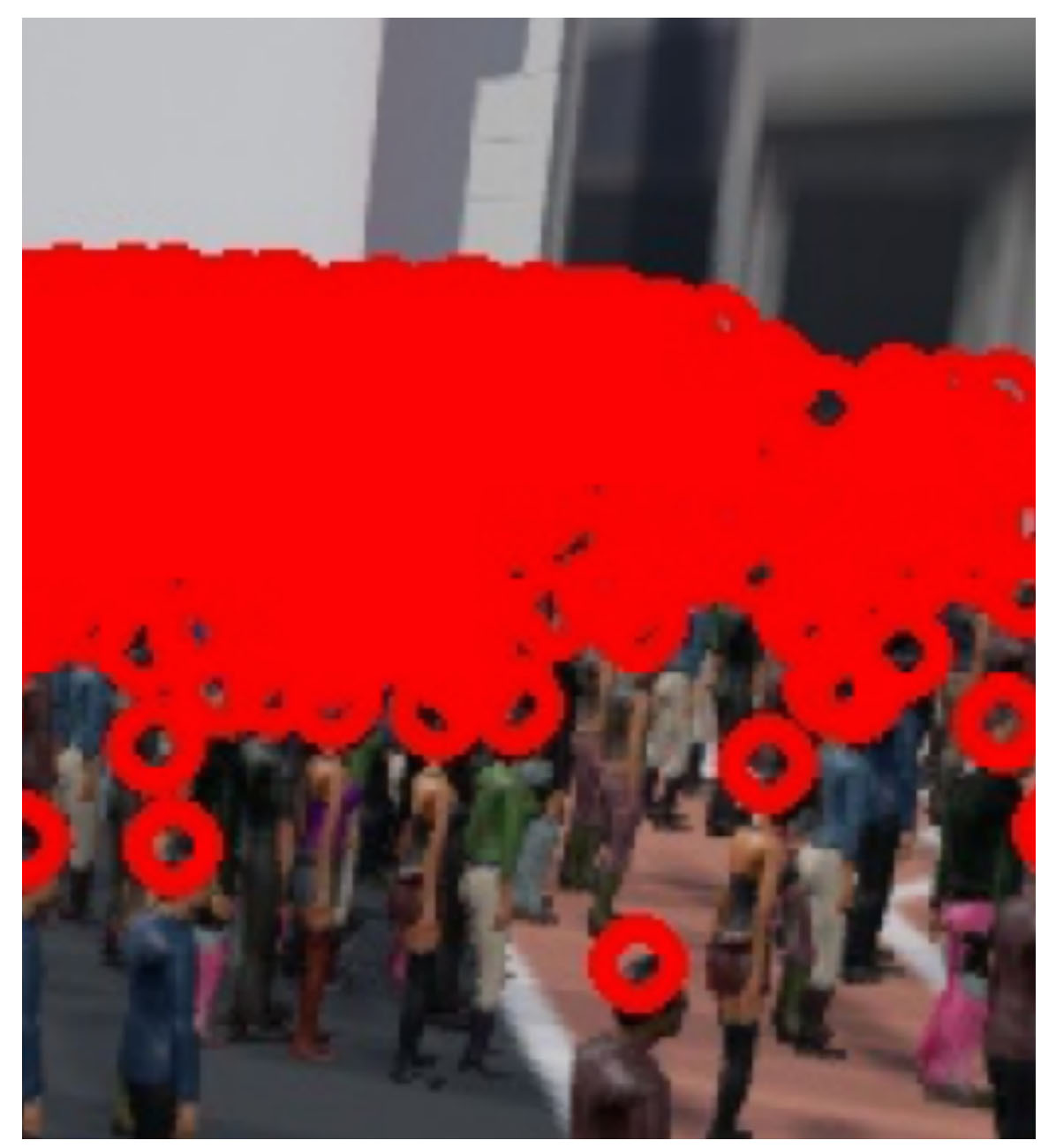}}~
\subfloat[]{\includegraphics[width=0.33\textwidth]{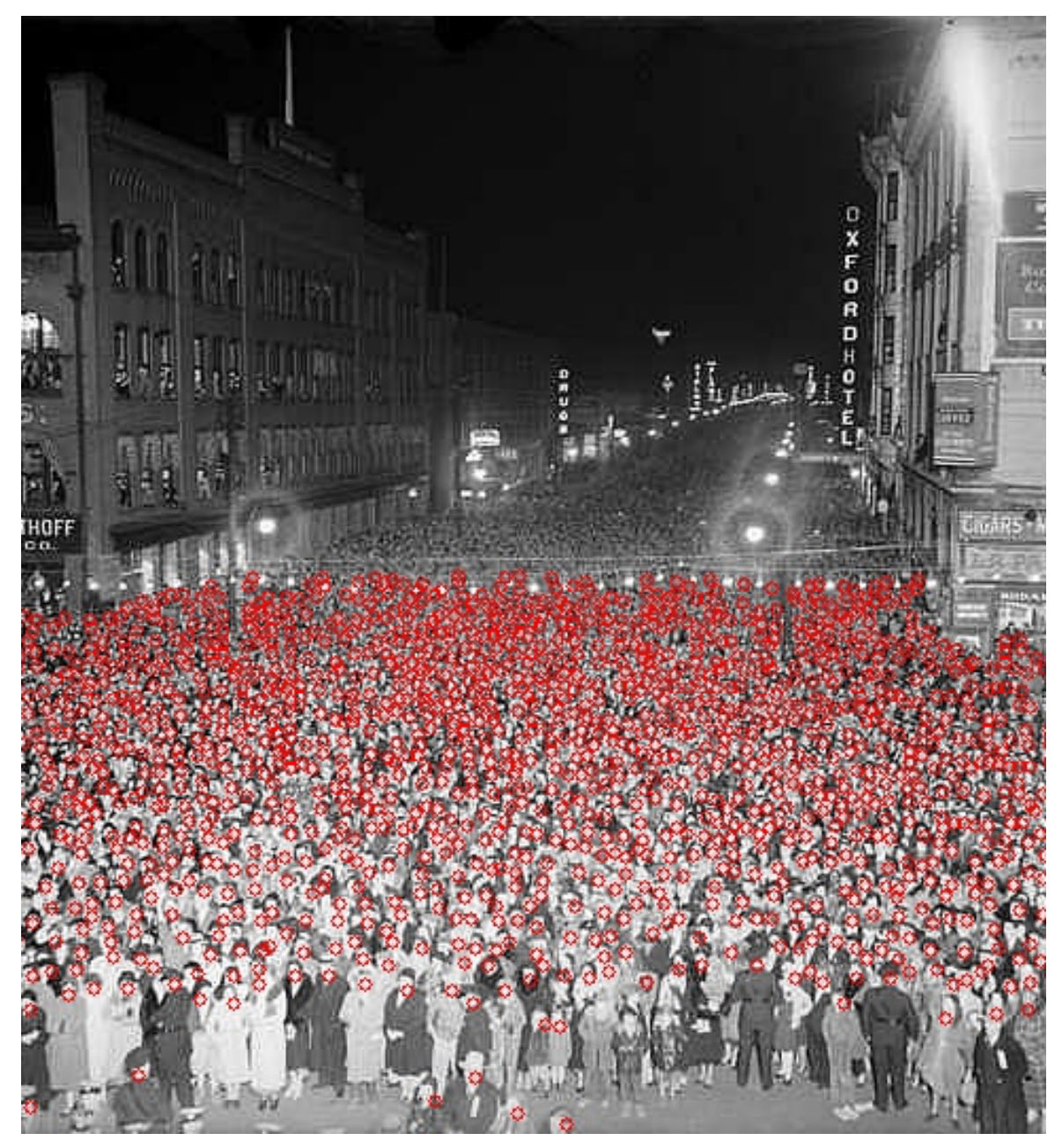}}
\vspace{-1em}
\caption{\small{(a) Generated using LCrowdV and consists of $858$ agents. (c) From UCF dataset~\cite{AliFlow}, it is very hard to accurately count the number of pedestrians or classify other characteristics such as density or behavior in this real-world image (i.e. generate accurate labels). In contrast, our method can automatically generate accurate labels as demonstrated in (b).}}
    \label{fig:crowd_photo}
    \vspace{-0.6cm}
\end{figure}

Machine learning methods, including deep learning, usually require a large set of labeled data to avoid over-fitting and to compute accurate results. A large fraction of crowd videos available on the Internet are not labeled or do not have ground truth or accurate information about the features. There are many challenges that arise in terms of using these Internet crowd videos for scene understanding:
\begin{itemize}
\item The process of labeling the videos is manual and can be very time consuming.
\item There may not be a sufficient number of videos available for certain crowd behaviors (e.g., panic evaluation from a large building or stadium) or for certain cultures (e.g., crowd gatherings in remote villages in the underdeveloped world). Most Internet-based videos are limited to popular locations or events.
\item The classification process is subject to the socio-cultural background of the human observers and their intrinsic biases. This can result in inconsistent labels for similar behaviors.

%For example, the pedestrians observed in the crowd in the Western world tend to have an expectation of larger personal spaces~\cite{chatt09}, as compared to those observed in Asia. This can lead to different or incorrect classifications of the pedestrians in terms of labeling, e.g. being classified as aggressive.
\item In videos corresponding to medium and high density crowds, it is rather difficult to exactly count the number of pedestrians exactly or classify their behaviors or tracks. This complexity is highlighted in one of the sample images in the UCF Crowd counting dataset~\cite{AliFlow}, shown in Figure~\ref{fig:crowd_photo}. Similar problems can arise in noisy videos or the ones recorded in poor lighting conditions.
\end{itemize}

In this paper, we present a new approach to procedurally generate a very large number of labeled, synthetic crowd videos for crowded scene understanding. Our approach is motivated by prior use of 
synthetic datasets in computer vision for different applications, including pedestrian detection~\cite{marin2010learning,hattori2015learning}, 
recognizing articulated objects from a single image~\cite{dhome1993determination}, multi-view car detection~\cite{movshovitz20143d,pepik2012teaching}, 3D indoor scene understanding~\cite{satkin2012data}, etc. In some cases, models trained using synthetic datasets
can outperform models trained on real scene-specific data, when labeled real-world data is limited.

%Extracting features from crowd behaviours is an significant problem to solve in the community.  Examples of features includes, head counting, head tracking, flow rate estimation(finding out how many people cross a line in the camera) and crowd behaviour classification. The application of the results is widely applied to different research fields and industries:  human-robot interaction, Social behaviour studies, surveillance and monitoring, etc.  

%A main stream of solution to the problem is developed by learning the features.   Labeling the data is possible, however it is very time consuming when the data set grows large. Currently, researchers in the community may heir a labour to label the ground truth, or make use of crowd-sourcing platforms.  Notwithstanding these ways can obtain labeled data, it is costly and it can only provide data up to a certain magnitude. 

\begin{figure*}[t]

    \centering
    \includegraphics[width=0.9\textwidth]{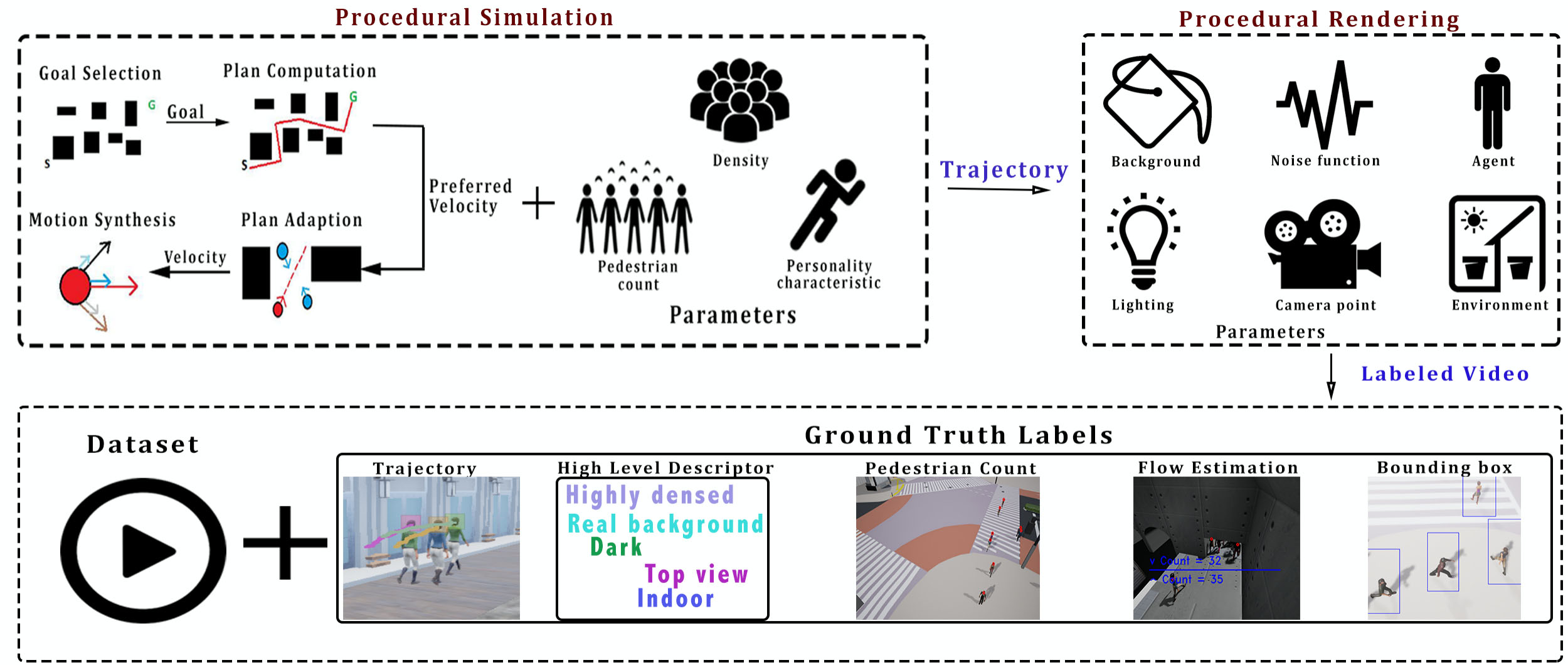}
    \caption{\small{LCrowdV framework consists of two components: procedural simulation (top left) and procedural renderer (top right). Each final video/image consists of a number of ground truth labels (bottom). Our approach can automatically generate different videos with accurate labels.}}
    \label{fig:flow}
    
    \vspace{-0.3cm}
\end{figure*}

\noindent{\bf Main Results:} We present a novel procedural framework to generate labeled crowd videos (LCrowdV). 
%Our approach is inspired by the use of procedural modeling methods in computer graphics and simulation.
% that are widely used to generate 3D models and textures~\cite{ebert2003texturing}. 
Our approach consists of two main components: procedural simulation and procedural rendering. We initially present a classification of crowd videos that is based on extensive research in sociology and psychology on crowd behaviors and crowd simulation. These classification criteria or parameters are used as the underlying labels for a given crowd video. Given a set of parameters or labels, our procedural framework can automatically generate an arbitrary number of crowd videos that can be classified accurately using those labels. These labels correspond to different indoor or outdoor environments, number of pedestrians, pedestrian density, crowd behaviors, lighting conditions, noise, camera location, abnormal behaviors, etc.
%Our approach can be extended and other crowd simulation or rendering methods can be easily incorporated into our framework. 

Our approach can be used to generate an arbitrary number of crowd videos or images (e.g. millions) by varying these classification parameters. 
%We can easily change the pedestrian density, pedestrian count, behaviors, movement, or any aspects of the environment to generate a new set of videos. 
Furthermore, we automatically generate a large number of labels for each image or video. The quality of each video, in terms of noise and resolution, can also be controlled using our procedural renderer. The generation of each video frame only takes a few milliseconds on a single CPU core, and the entire generation process can be easily parallelized on a large number of cores or servers. We included a small subset of LCrowdV in the supplementary material, though the entire data would be made available on the WWW.
 
We demonstrate the benefits of LCrowdV over prior labeled crowd video datasets on the following applications:
\begin{itemize}
\item {\bf Improved Pedestrian Detection using HOG+SVM:} We demonstrate that combining LCrowdV with a few real world annotated videos can improve the average precision by $3\%$. In particular, the variations in the camera angle in the LCrowdV dataset have the maximal impact on the accuracy. Instead of $70$K labeled real-world videos, we only use $1$K annotated real-world videos along with LCrowdV to get improved accuracy.
\item {\bf Improved Pedestrian Detection using Faster R-CNN:} We demonstrate that combining LCrowdV with a few real world annotated videos can improve the average precision by $7.3\%$. Furthermore, we only use $50$ labeled samples from the real-world dataset and combine them with LCrowdV.
\item {\bf Improved Crowd Behavior Classification:} We use Eysenck Three factor personality model in psychology to generate a high variety of behaviors in LCrowdV. As a result, we can improve the behavior classification in real-world videos by $9\%$ up to $41\%$.
\end{itemize}

%crowd behavior analysis based on personality models and pedestrian detection. We use a well known linear regression based on Eysenck Three factor personality model in psychology, and the simulation parameters in an input video that are computed using an online pedestrian tracker. We use the large LCrowdV videos to classify the personality of each pedestrian in the input video
%We also trained a HOG+SVM pedestrian detector with our simulated data to show that our label dataset can generate comparable (or better) results to a descriptor trained by real-world crowd datasets.
 
 The rest of the paper is organized as follows. We give a brief overview of prior work on crowd analysis and crowd datasets in Section 2. We describe our procedural framework in Section 3 and highlight its benefits for pedestrian detection and crowd behavior classification in Section 4.

\section{Related Work}

The simulation, observation and analysis of crowd behaviors have been extensively studied in computer graphics and animation, social sciences, robotics, computer vision, and pedestrian dynamics~\cite{ali2013modeling}.  In this section, we give a brief overview of recent work on crowd video analysis, classification, labeling, and prior crowd datasets.
%\textcolor{blue}{ Note that our work utilizes the prior work on crowd simulation and validation. Our main novelty is in terms of designing a novel algorithm and system for procedural simulation that can automatically generate a very large collection of crowd behaviors (e.g. 1M or even 10M samples) automatically.}

%\begin{itemize}
%  \color{OliveGreen}
%  \item OTHER CROWD AND PEDESTRIAN DATASETS (e.g. UCF WORK OR OTHERS CITED IN XIAOGANG'S WORK) YOU ALSO NEED %TO MENTION HOW BIG OR COMPLEX WERE THOSE DATASETS, WHAT BENEFITS OR FEATURES THEY SUPPORTED and WHAT DID %THEY LACK
%\end{itemize}

%\textcolor{OliveGreen}{
%WHAT ARE THE MAXIMUM CAPABILTIES OF THESE SYSTEMS? YOU WANT}

\subsection{Crowd Video Analysis}
Different models have been proposed to model the crowd behaviors \cite{zhou2012understanding,hospedales2009markov,Mehran_NormalAbnormalSocialForce,4359353,4731265,bera2016realtime,kim2015interactive}.  Other methods focus on extracting specific features in crowds, including  head counting~\cite{rabaud2006counting,5540833,1683826,garcia2009density,idrees2013multi,zhang2015cross}. Online trackers  tend to use the current, previous or recent frames for tracking each pedestrian in the video at interactive rates
% . These trackers are based on non-adaptive random projections that model the structure of the image feature space of objects
~\cite{zhang2012real,Fu2012,song2013fully,AliFlow}.
% or learning semantic motion patterns for dynamic scenes by improved coding~\cite{Fu2012}. Other tracking algorithms use pedestrian motion features to compute the trajectory of the agents. These algorithms include clustering methods based on the assumption that pedestrians only appear and/or disappear at entry and/or exit points~\cite{song2013fully}, and flow-field based methods can determine the probability of motion in densely crowded scenes~\cite{AliFlow}.
Tracking the pedestrians to obtain the full trajectories has been extensively studied ~\cite{mikel_dataDriven,zhou2012understanding,ali2008floor,zhu2014crowd}. 
There is considerable research on analyzing various crowd behaviors and movements from videos~\cite{LiCrowdedSceneAnalysis2015}. 
%The main goals of these methods include extracting useful information regarding behavior patterns, performing crowd activity recognition or detecting abnormal behavior or situations for surveillance analysis.
%Certain methods focus on classifying the most common behavior patterns in a given scene. 
Most of these methods are designed for offline applications and tend to use a large number of training videos to learn the patterns~\cite{solmaz2012identifying,Mehran_NormalAbnormalSocialForce,kratz2009anomaly,mikel_dataDriven,hospedales2009markov}.
% for the following applications: detecting common crowd behavior patterns~\cite{Shah_crowdPatterns}, detecting normal and abnormal interactions~\cite{Mehran_NormalAbnormalSocialForce,kratz2009anomaly}, or detecting human group activities~\cite{Ni2009}; as well developing approaches that use a large selection of videos on the web as examples of certain types of motion~\cite{mikel_dataDriven}. \cite{hospedales2009markov, Mehran_NormalAbnormalSocialForce, 1698861} detect abnormal behaviors in crowd videos.  
Different methods have also been proposed to estimate the crowd flow in videos~\cite{6027295,chen2006intelligent,tsuduki2009method,5597451} , model activities and interactions in crowded and complicated scenes \cite{4359353,4731265,5206827}.

Many crowd analysis methods \cite{hospedales2009markov,Mehran_NormalAbnormalSocialForce,6027295,zhou2012understanding,4359353,1698861,4731265,5206827} tend to be scene specific, which implies that they are trained from a specific scene and haven't been tested in terms of generalizing the results across other scenes.  One of the challenges is to find complete crowd datasets that include data samples covering enough scenes and behaviors, and provide labeled ground truth data for learning.  Some methods \cite{solmaz2012identifying,5540833,chen2006intelligent} don't require real data for training, but they are limited by the size of the crowds or specific conditions, including crowd behaviors and color information.

\subsection{Crowd Classification and Labeling}
Crowd behaviors are diverse, and it is a major challenge to model different behaviors. \cite{solmaz2012identifying} classified crowd behaviors in five categories in accordance with their dynamical behavior: bottlenecks, fountainheads, lane formation, ring/arch formation and blocking.  \cite{5653573} classified crowd behaviors into another five categories based on its dynamics:  running, loitering, dispersal(center to edge), dispersal(edge to center) and formation.  Interestingly, these two methods cannot cover all behaviors  that are observed in crowd videos.  
%There are other methods that do not emphasize classifying every crowds into a set of categories, but they instead 
Other methods focus on labeling the crowd data that can be described by a predefined set of labels\cite{rodriguez2008action,reddy2013recognizing,DBLP:journals/corr/abs-1212-0402}. Recently, \cite{shao2015deeply} uses a model along with manually entered labels to classify crowds based on the location, the subject and the action.  Our approach is motivated by these prior works on crowd classification and labeling.
%In this dataset we tend to use the labeling approach instead of the classifying approach, as we are aware of the difficulty in defining a set of classes that covers all the crowd videos. 

\begin{table*}[t]
  \scriptsize
  \centering
  %\begin{tabular}{|p{2cm}|p{1cm}|p{1cm}|p{1cm}|p{1cm}|p{1cm}|p{1cm}|p{1cm}|p{1cm}|}
  \begin{tabular}{|c|c|c|c|c|c|c|c|c|}
  \hline
    Dataset & \specialcell{CUHK \\ \cite{6909682}} & \specialcell{Collectiv-\\eness \cite{6714561}}  & \specialcell{Violence \\ \cite{6239348}} & \specialcell{Data- \\Driven \cite{mikel_dataDriven}} & \specialcell{UCF \\ \cite{AliFlow}} & \specialcell{WWW \\  \cite{shao2015deeply}} & \specialcell{CVC07 \\ \cite{xu:2014}} & \textbf{LCrowdV} \\ \hline
    Videos & 474 & 413 & 246 & 212 & 46 & 10000 & N/A & \textbf{\textgreater 1 M} \\ \hline
    Frames & 60,384 & 40,796 & 22,074 & 121,626 & 18,196 &\textgreater 8 M & 2,534 & \textbf{\textgreater 20 M} \\ \hline
    Resolution & Varying & 670x1000 & 320x240 & 720x480 & Varying & 640x360 & Varying & \textbf{Any} \\ \hline
 %   Type of ground truth &  \multicolumn{8}{|c|}{ } \\ \hline
    Trajectory &x&x&x& \checkmark &x&x&x& \checkmark \\ \hline
    Pedestrian count &x&x&x&x&x&x&x& \checkmark \\ \hline
    Flow estimation &x&x&x&x&\checkmark&x&x& \checkmark \\ \hline
    Attributes & 3 & 1 & 0 & 2 & 0 & 3 & 0 &7 \\ \hline
    Bounding box &x&x&x&x&x&x& \checkmark & \checkmark \\ \hline
    Generation Method & \multicolumn{6}{|c|}{Manually} & \multicolumn{2}{|c|}{\textbf{Automatically}} \\ \hline

  \end{tabular}
  \caption{\small{{\bf Benefits of LCrowdV:}  Not only can we generate a significantly higher (or arbitrary) number of labeled videos, but we can also provide a large set of crowd labels and characteristics for each image and  video. We can also control the behavior characteristics, environments, resolution and rendering quality of each video to develop a good training set. Unlike prior methods, we can easily generate accurate labels.}}
  \label{tab:param}
  
  \vspace{-0.5cm}
\end{table*}

\subsection{Crowd Video Datasets}
Many crowd video datasets that are available can provide ground truth or estimated labels for analysis or training.  \cite{mikel_dataDriven} provides trajectories of the pedestrians,\cite{rabaud2006counting} describes a database with number of pedestrians for crowd counting,  \cite{hattori2015learning} includes bounding boxes of the detected pedestrians, and  \cite{5597451} provides ground truth labels for crowd flow estimation.   Shao et al. \cite{shao2015deeply} consists of high-level labels to describe crowd characteristics.  Most of these datasets are generated by labeling the data manually, and this process is time consuming and error prone.  For example, to allow the deep learning model in \cite{shao2015deeply} to understand the scenes and put appropriate labels, $19$ human annotators were involved. Several human operators were needed for determining the number of objects in \cite{rabaud2006counting}.  Table \ref{tab:param} shows the comparison of existing crowd datasets, as compared to LCrowdV.

\subsection{Learning Crowd Behaviors with Simulated Data}
\cite{marin2010learning} trained a pedestrian detector for a driver assistant system using simulated data.  This work focused on the training and testing data from one particular camera angle, which corresponds to the driver's view.  \cite{hattori2015learning} generated simulated agents on a specific real-world background to enable  learning for pedestrian detectors.  In contrast with these methods, our work aims at providing a diversified, generic and comprehensive approach for generating different crowd video data for analyzing different crowd behaviors.

\section{Synthetic Label Crowd Video Generation}
Crowds are observed in different situations in the real-world, including indoor and outdoor scenarios. One of our goals is to develop a procedural framework that is capable of providing all types of crowd videos with appropriate ground truth labels. In this section, we give an overview of our framework and the various parameters used to generate the videos.

\begin{figure}
    \centering
    \includegraphics[width=0.9\textwidth]{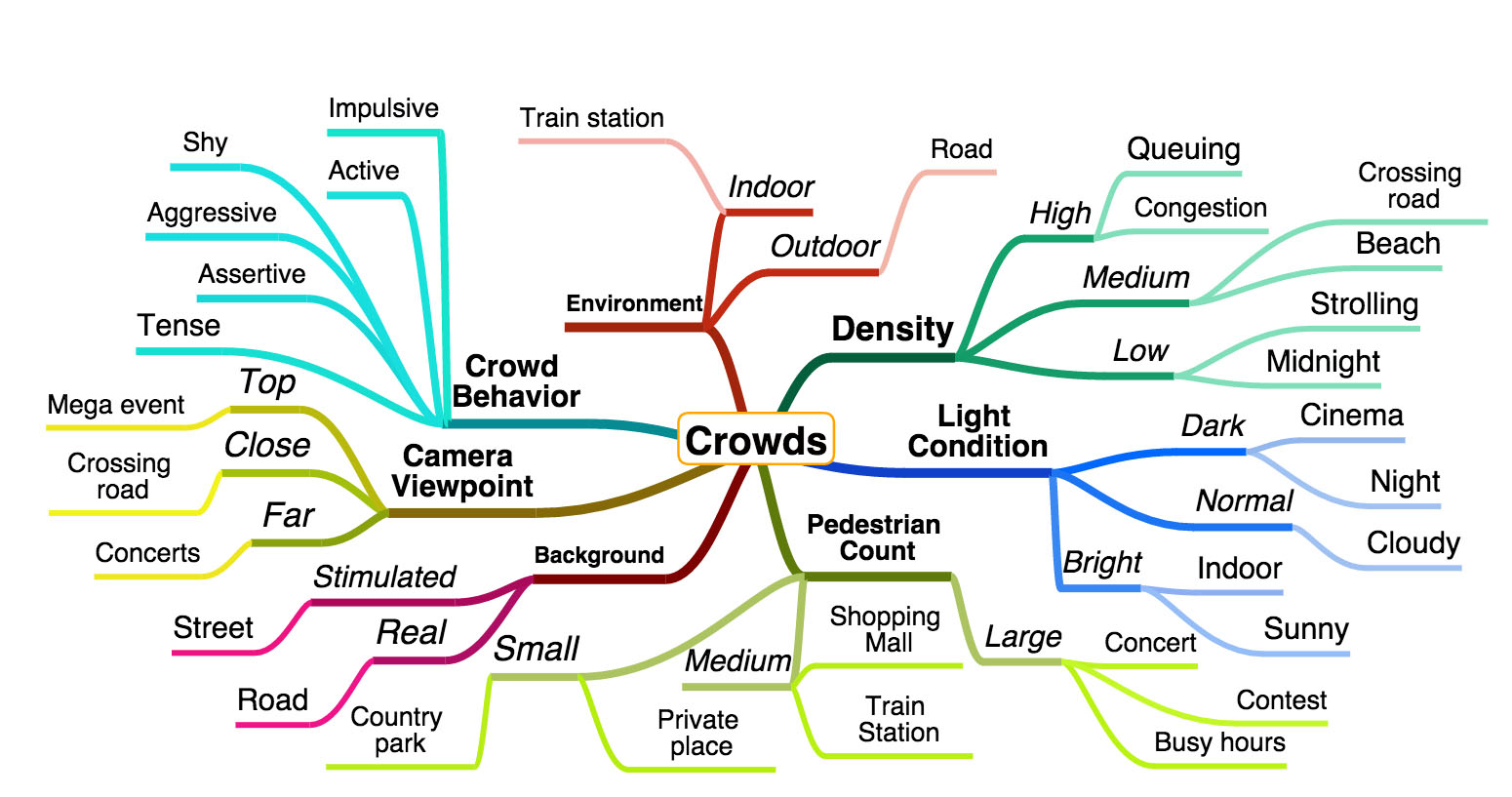}
    \caption{\small{Hierarchical and parametric classification of crowd behaviors and renderings.  Attribute Labels of LCrowdV includes:  Background, Crowd Behaviour, Camera Viewpoint, Density, Environment, Light Condition and Pedestrian Count.  We use these labels to classify different characteristics of crowds and use them in our procedural  framework.}}
    \label{fig:tree}
    \vspace{-0.7cm}
\end{figure}

%\textcolor{blue}{
%The outline of our proposed method is described in \ref{fig:flow}.  We first generate the trajectories of each individual pedestrians (agents) in the simulated world.  After that, we combine several elements with the agent trajectories to render a video.  At the same time, we produce a set of ground truth labels.
%}

\subsection {Crowd Classification and Generator}
Modeling the behavior of large, heterogeneous crowds is important in various domains including psychology, robotics, pedestrian dynamics, and computer graphics. There is more than a century of work in social sciences and psychology on observing and classifying crowds, starting from the pioneering work of Lebon~\cite{Lebon1895}. Other social scientists have classified the crowds in terms of behaviors, size, and distributions~\cite{james1953distribution}. According to {\em Convergence Theory}, crowd behavior is not a product of the crowd itself; rather it is carried into the crowd by the individuals~\cite{turner1987collective}. These observations and classifications have been used to simulate different crowd behaviors and flows ~\cite{pervin2003science,moussaid2010walking,eysenck1985personality}. 

Our procedural crowd simulation framework builds on these prior observations in social sciences and on simulation methods. The overall hierarchical classification is shown in Fig.~\ref{fig:tree}. Each of these labels is used to describe some attributes of the crowds or the pedestrians. In addition to the labels that govern the movements or trajectories of each agent or pedestrian, we also use a few labels that control the final rendering (e.g., lighting, camera position, brightness, field of view) of the crowd video by using appropriate techniques from computer graphics.
%In particular, we use standard rendering techniques from computer graphics to generate each image or frame of the crowd. These rendering parameters are used to control the  camera position and field of view, the scene lighting conditions, as well as other environmental components. 
Finally, we can also add some noise (e.g. Gaussian noise) to degrade the final rendered images, so that these images are similar to those captured using video cameras.

\noindent{\bf Framework Design:} Our framework has two major components: procedural simulation and procedural rendering. The procedural simulator takes as input the number of agents or pedestrians, densities, behavior and flow characteristics and computes the appropriate trajectories and movements of each agent corresponding to different frames.   Given the position of each agent during each frame, the procedural renderer generates the image frames based on the different parameters that control the lighting conditions. Each of these input parameters corresponds to a label of the final video. 

\noindent{\bf Procedural Crowd Simulation:}
In this section, we give an overview of our procedural crowd simulator.  We use the Menge crowd simulation engine~\cite{curtis2016menge}, which makes it possible to combine state-of-the-art crowd modeling techniques addressed in the previous section.  Given the labels or high-level descriptors, our method can generate crowd movements or behaviors that fit those descriptions. These include the total number of agents or pedestrians in the scene, their density over different parts of the environment or the scene, their global and local movements, and the behavior characteristics. 

\noindent{\bf Global Crowd Characteristics:}
 %EXPLAIN HOW THESE PARAMETERS ARE USED TO CHARACTERIZE THE CROWD BEHAVIORS; CALL THEM GLOBAL PARAMETERS. HOW ARE THESE PRAMETERS SPECIFIED? WAHT ARE THE DIFFERENT PERSONALITY MODELS THAT CAN BE USED TO CHARACTERIZE EACH AGENT
In the simulation stage, we vary the global parameters, including the personality settings of different agents, density, and the number of agents used to generate different types of trajectories.  The number of agents will control how many pedestrians are in the scene, and the density factor decides whether or not the pedestrians would be located very close to each other.  
%These two parameters are important as they affect how much overlapping between the agents will be there in the final rendered image.
  It is essential to include different levels of overlapping in the training data set to avoid over-fitting.  The personality parameters allow the crowd behavior to be more natural-looking.    

\noindent{\bf Crowd Movement and Trajectory Generation:}
A key component is simulating the crowd or pedestrian movement. We build on prior research in crowd movement simulation~\cite{funge99,ulicny2002towards} and use the property that movement specification can be decomposed into three major subproblems: agent goal selection, global plan computation, and local plan adaptation
 (see Fig. \ref{fig:flow}).  We further elaborate on each of these components and give various possible solutions for each of these subproblems, based on the different crowd behavior classifier.   
 
\noindent{\bf Goal Selection:}
In the goal selection module, we specify the high-level goal of each pedestrian. For example, the agent may want to go to a particular location in the scene or just visit a few areas, etc. It is expected that the goal can change across time and is affected by the intrinsic personalities or characteristics of the agents or the environmental factors. There is extensive literature on goal selection methods and we can use these methods in our procedural simulation framework~ \cite{shao2005autonomous,ulicny2002towards}. 

\noindent{\bf Global Path Planning:}
Given the goal specification of each agent, we compute collision-free trajectory to achieve that goal. 
%This depends on whether the goal requires an agent to perform an action without locomotion or traversing to another position.
% The problem will be deferred to the full human motion computation (see below) or is solved using path planning methods.  In our case, t
The path-planning module is a high-level module that computes a preferred velocity or direction for each agent for a given time-step. We use techniques based on potential field methods, roadmaps, navigation meshes, and corridor methods~  \cite{barraquand1991robot,snook00,lamarche2004crowd,geraerts2008using}.

\noindent{\bf Local Plan Adaptation:}
Since the path computed in the previous stage usually considers only the static obstacles, we need to adapt the plan to tackle the dynamic obstacles or other agents or pedestrians in the scene.   We transform the preferred velocity computed into a feasible velocity that can avoid collisions in real time with multiple agents.
% This local plan adaptation step is computed in a distributed manner for each agent with the assumption that a consistent solution used by all agents would guarantee collision avoidance.
Some of the commonly used motion models for local plan adaptation are based on social forces, rule-based methods, and reciprocal velocity obstacles~\cite{boids,helbing1995social,Jur:2011:RVO}.

\noindent{\bf Full Human Motion Computation:}
The main goal of high DOF human motion computation or motion synthesis is to compute the locomotion or position of each agent in terms of the joint positions, corresponding to the walk cycle as well as to the motion of the head and upper body. We use standard techniques from computer animation based on kinematic, dynamics and control-based methods to generate the exact position of each pedestrian in the scene~\cite{brude89,Lee07,vanB11}.

\subsection {Procedural Rendering:}
After computing the trajectory or movement specification characterized by the global parameters for each pedestrian in the video, we generate an animation and render it using different parameters. We can control the lighting conditions, resolution, viewpoint, and the noise function to lower the image quality, as needed.

\noindent{\bf Animated Agent Models:}
We use a set of animated agent models, that include the gender, skin color, clothes and outlook.  We randomly assign these models to every agent.  Furthermore, we may associate some objects in the scene with each agent or pedestrian. For example, in the case of a shopping mall, a customer may carry items he or she bought in bags;  and in a theme park, there may be parents walking along with the children. These attached items could potentially obstruct the agent and change its appearance.

\noindent{\bf 3D Environments and Backgrounds:}
Our background scenes include both indoor and outdoor environments. Ideally, we can import any scene with a polygonal model representation and/or textures and use that to represent the environment. We can also vary the lighting conditions to model different weather conditions:  a sunny day could have a huge difference in appearance compared to a gloomy day.  On top of that, we can also add static and dynamic obstacles to model real world situations.  For instance, we could add moving vehicles into a city map and animated trees into a countryside map.  
%These obstacles are different from attached items because they are either static or they have independent trajectories verses the attached items that are moving relative to an agent.

\noindent{\bf Image-space Projection and and Noise Functions:}
In order to render the 3D virtual world and the animated agent model, we render the image using a camera projection model: perspective projection or orthogonal projection.  Typically, we render the videos with perspective projections to simulate real world videos.  At this stage, we use different parameters to the projection model to obtain the videos captured from different viewpoints. In practice, video and images collected from different camera views could result in significant differences in the appearance.  We also add a Gaussian noise filter with varying levels of standard deviation to emulate the digital noise in a video frame.

In our current implementation, we use the Unreal game  engine\cite{Oliver:2012:UEE:2341836.2341909} for rendering and generating each video frame. It is an open source engine and we can easily specify the geometric representation, lighting and rendering information, and generate images/videos at any resolution or add noise functions. We can easily use different rendering parameters into Unreal Engine to control the final crowd rendering.

%, which is a large-scale game engine.  The fact that Unreal Engine is open source allows us to develop it for our rendering procedure.  

%As for the results generated, we currently provide four sets of ground truth labels along with the videos:  Head trajectories, bounding boxes of pedestrians, crowd flow count, and high-level descriptors.  

%KEEP THIS SMALL
\subsection{Ground Truth Labels}
The two main labels related to such datasets including the pedestrian count as well as the trajectory of each pedestrian.
%Two of the major problems in such datasets are related to crowd counting and accurately tracking the trajectory of each pedestrian. 
This can be rather challenging for dense crowds, where generating such a labeled dataset is a major challenge.  In order to accurately generate such labels,  we consider each head of an agent in the video that is not obstructed by other scene objects. We compute the screen-space coordinates during each frame for every agent using the given camera parameters.  We can also compute the position of lower body or full body contours. Given these head and lower body information, we can accurately compute the count and the trajectories.
%The screen coordinate is directly computed from the clipping plane coordinates $[ n, f,  t,  b,  l,  r ] ^ {T}$ and the head coordinate of the agent in the simulated world $[X_c, Y_c, Z_c]$:
%\begin{center}
% $\begin{pmatrix} 
% u_i \\ v_i \\ z_i \\ 1
% \end{pmatrix}
% =
% \begin{pmatrix}
%  \frac{2n}{r-l} & 0 & \frac{r+l}{r-l} & 0 \\
%  0 & \frac{2n}{t-b} & \frac{t+b}{t-b} & 0 \\
%  0 & 0 & \frac{-(f+n)}{f-n} & \frac{-2fn}{f-n} \\
%  0 & 0 & -1 & 0 %
% \end{pmatrix}
% \begin{pmatrix}
% X_c \\ Y_c \\ Z_c \\ 1
% \end{pmatrix}$
%\end{center}
%Therefore, the screen coordinate $(u_i, v_i)$ obtained through this way directly is more precise (up to sub-pixel accuracy) than the existing method of labeling the heads % manually, which involves human error.  Also, it is fully automatic and does not involve the gigantic labour effort to perform the label task. 
%\subsubsection{Ground Truth Labels: Bounding Boxes of Pedestrians}

Apart from the trajectories of the head, we also use the bounding boxes for pedestrian detection.  Using the same technique mentioned above, we compute the bounding box for each pedestrian, which is centered at the centroid of the model used for each agent.  This is  more accurate than annotating the bounding boxes manually, especially for high density scenes.
%\textcolor{OliveGreen}{
%// TODO:  Add saamples of boudning boxes, trajectories, and crowd flow count
%}
%WHAT IS THE GOAL? IS THIS A HIGH LEVEL PARAMETER?
%\subsubsection{Ground Truth Labels: Crowd Flows}

Another major problem in crowd scene analysis is computing the crowd flows.  The goal is to estimate the number of pedestrians crossing a particular region in the video.  For real videos with dense pedestrian flows, it is difficult and labor intensive to compute such flow measures.  This is due to the fact that there may be partial occlusion and a human operator needs to review each frame multiple times  to obtain this information accurately. 
On the other hand, we can easily count how many agents are crossing a line or a location in the scene. However, in some of the pedestrian videos, agents could walk around or over the counting line because of collision avoidance.  If we can count every agent that crosses the line, this count could increase when an agent is close to the counting line or when an agent repeatedly crosses the line.  Therefore, we define an agent as crossing the line only if it has passed a particular tolerance zone or region in the scene. 

\begin{figure*}

    \vspace{-0.1cm}
    \centering
    \includegraphics[width=0.9\textwidth]{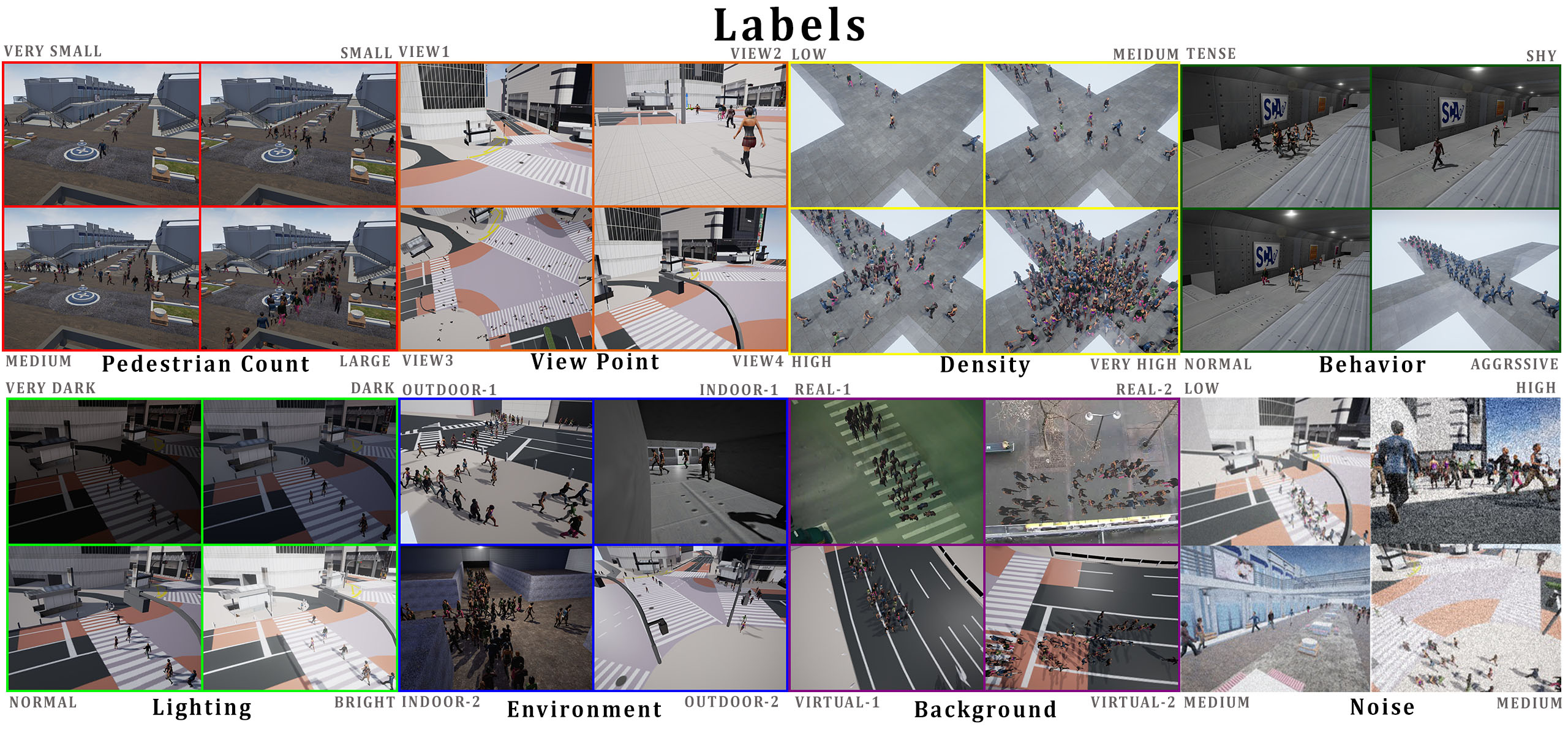}
    \caption{\small{{\bf Parameters used in LCrowdV}: Samples of images that illustrate the effect of changing different high level parameters in the scene.  These parameters are also used as the labels.}}
    \label{fig:label}
    \vspace{-0.7cm}
\end{figure*}

In addition to these labels corresponding to the tracks, bounding boxes, flows, etc. we also keep track of all the parameters used by the procedural framework, i.e. the seven different high level parameters. As mentioned in the previous section, we have generated the videos using seven different parameters.  These parameters can be used to describe the video in a high level manner, as shown in Fig. \ref{fig:tree} and Fig. \ref{fig:label}.

\section {Applications and Evaluation}
In this section, we highlight the benefits of LCrowdV dataset for improved crowd behavior classification and pedestrian detection.

\noindent {\bf Crowd Datasets:} For our evaluations, we used many real-world datasets: CUHK Crowd dataset~\cite{zhu2014crowd}, INRIA\cite{1467360}, KITTI\cite{6248074}, ETHZ\cite{4587581}, and Town Center\cite{benfold2011stable}. 
The INRIA dataset contains $1832$ training and $741$ testing sample images.
We used the object detection dataset in KITTI Vision Benchmark Suite.  As annotations are not provided in the test set, we divided the train set into two subsets: 1279 images for training and 500 images for verification.  
In the ETHZ dataset, trained with BAHNHOF (999 image frames) and JELMOLI (936 image frames) and tested on SUNNY DAY (354 frames).
The Town Center dataset is a 5-minute video with 7500 frames annotated, which is divided into 6500 for training and 1000 for testing data for pedestrian detection.
%MOVE THIS DESCRIPTION ABOVE.  >?>>>>> like this?  move before INRIA?
 We have created a new dataset called Person Search Database(PSDB). This database consists of $18,184$ images taken at different angles. And unlike the Town Center dataset which consist of images at the same viewpoint and scene, the scenes in PSDB are more diverse, including shopping mall, roadside, University, park, etc. 
For behavior analysis, we evaluated the crowd motion trajectories of the pedestrians, as opposed to the actual appearance. For pedestrian  detection, we use selected frames from the dataset.

\subsection{Crowd behavior Classification}
Our goal is to automatically classify a crowd video based on the collective personality of the individuals or pedestrians. In particular, each individual may be characterized as shy, aggressive, or tensed based on its movement pattern. In our formulation, we use the well known personality trait theory from psychology and  use the Eysenck 3-factor model~\cite{eysenck1985personality} to classify such behaviors. This model identifies three major factors that are used to characterize the personality: Psychoticism, Extraversion and Neuroticism (commonly referred to as PEN). Each individual personality is identified based on how they exhibit each of these three traits. These individual personalities are combined and the overall crowd behavior can classified into six behavior classes - aggressive, assertive, shy, active, tense and impulsive. 
%To our knowledge there is no such prior work on classifying crowd behaviors based on this 3-factor model.

\begin{figure}

\vspace{-0.2cm}
\centering
\includegraphics[width=0.8\linewidth]{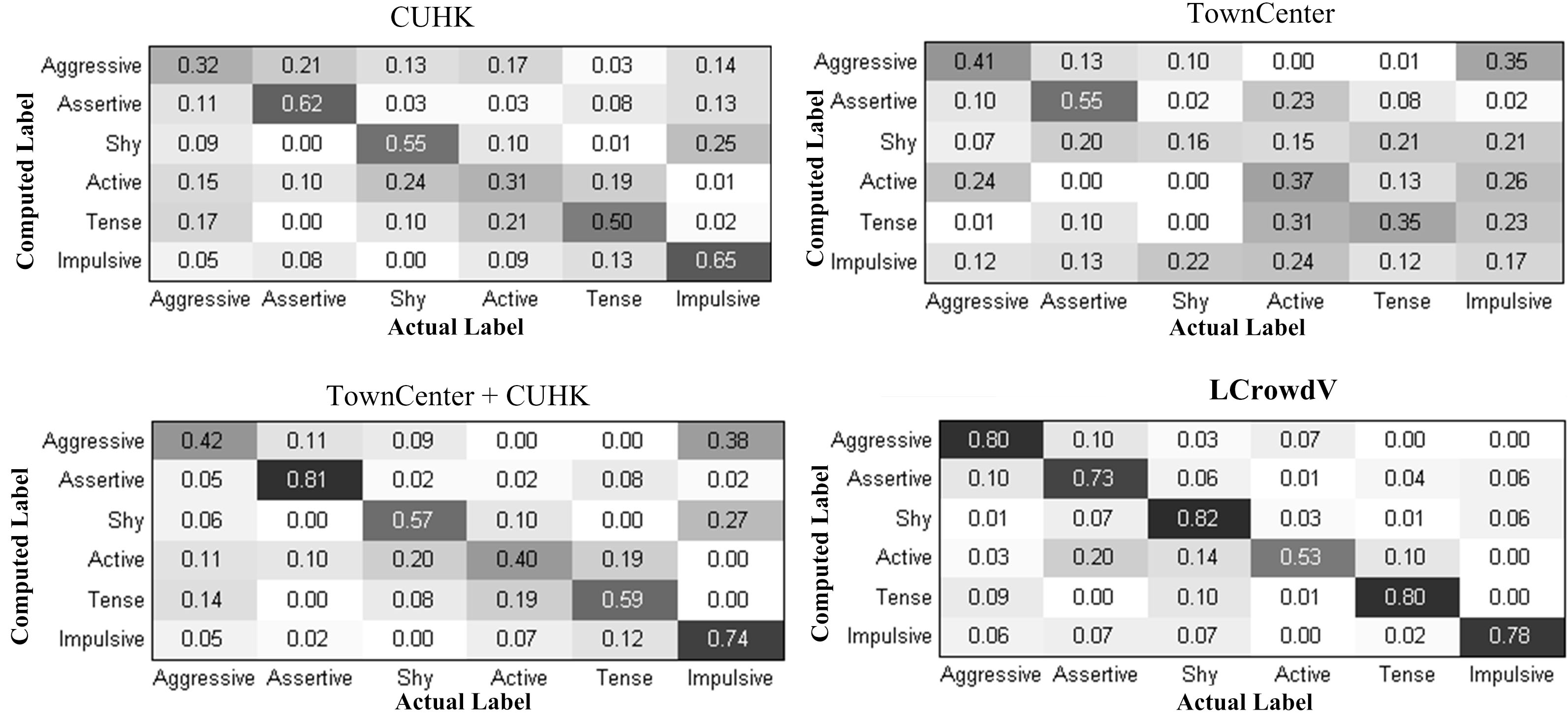}\vspace{-1em}
\caption{\small{\textbf{Confidence Matrix:} The X-axis shows the actual behavior label in the original video and Y-axis represents the computed label used by classification algorithm. The scale is from 0 to 1. For example, a cell with 0.3 with 'Shy' X-axis and 'Assertive' Y-axis indicates that 30\% of the actual 'Shy' videos have been classified as 'Assertive' by the classification system. We observed that using LCrowdV for training significantly increases the accuracy in terms of correct behavior classification, even more than 'Town Center' and 'CUHK' combined.} }\vspace{-1em}
\label{fig:matrix}

\end{figure}

Our behavior classification algorithm is based on computing a linear regression between the crowd simulation parameters and the perceived behaviors of each agent based on the  Eysenck  model. The input to the regression formulation is the difference between the given agents' parameters, used by the underlying multi-agent simulation algorithm (Section 3.2), and those computed in the reference video using the online tracker.  This formulation removes the need to compute any offset as part of the regression. 
\begin{figure}[t]
\centering
\includegraphics[width=0.9\linewidth]{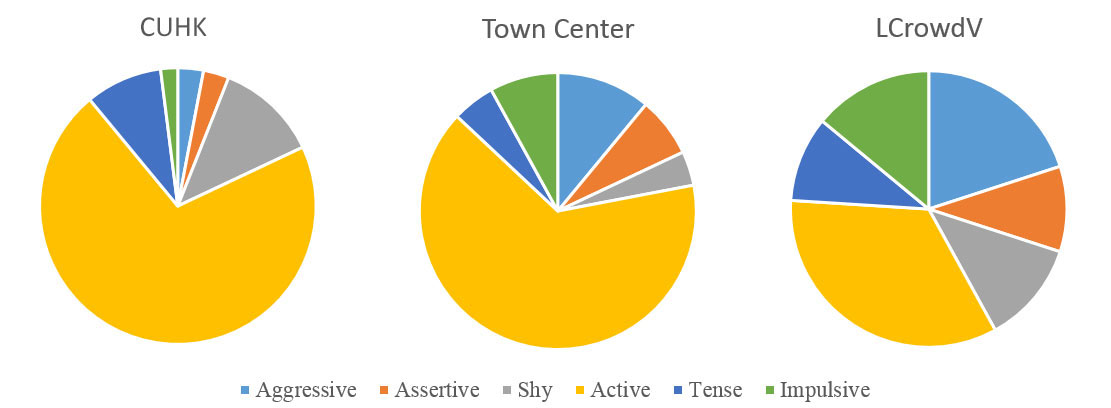}\vspace{-1em}
\caption{\small{\textbf{Crowd Behavior Distribution:} We observe that even after manually labelling prior realworld crowd datasets for behavior classification, LCrowdV includes a wider spectrum of videos and variations for every behavior classification. This highlights the benefit of LCrowdV in terms of automatically generating a wide variety of labeled videos.}}\vspace{-1em}
\label{fig:variance}
\end{figure}

After computing the best-fit parameters, we perform a lookup to find the best match in the  table in LCrowdV using  kd-tree nearest neighbor feature matching. 
%Our mapping for the crowd behavior classification is derived a the user-study.
For more details regarding the specific parameters used in our mapping, please  refer to Section 1 of the Appendix in the supplementary materials.

\begin{figure}[t]
    \centering
    \includegraphics[width=1\textwidth]{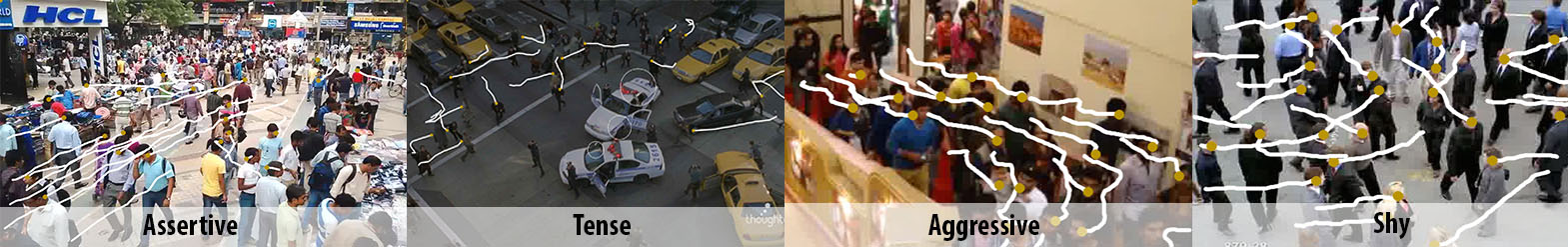}\vspace{-1em}
    \caption{\small{Sample output of our crowd behavior classification algorithm. The white lines represent the pedestrian trajectories computed using an online tracker. These trajectories are used to automatically compute the best fit personality trait of each pedestrian and classify its behavior. Although individuals have different personality models, we collectively classify the entire video clip with a single personality model.}
    % (Bottom) We tracked 37 different web-sourced videos and we categorized the different personality traits based on our crowd behavior classification.
    }
    \label{fig:behavior}
     \vspace{-0.4cm}
\end{figure}
We evaluate the accuracy of our behavior classification algorithm using  $50$ crowd sourced videos from YouTube. Since other crowd datasets (CUHK Crowd Dataset~\cite{zhu2014crowd}, TownCenter~\cite{benfold2011stable}, etc.,) do not have behavior annotation, it is very hard to use them for such evaluation. In order to perform a fair comparison, we manually labelled a subset of videos in CUHK and Town Center Datasets, a total of $48$ video clips, and compared the results with LCrowdV. We observe that the large collection and variance (Refer: Fig. ~\ref{fig:matrix} and ~\ref{fig:variance}) of the videos in LCrowdV provides more accurate results than prior realworld datasets. The Confidence Matrix in Fig. ~\ref{fig:matrix} clearly highlights the benefits of LCrowdV over CUHK and Town Center datasets. With LCrowdV, 80\% of the videos showing aggressive, shy, tense and impulsive behaviors have been classified correctly.

\subsection{Pedestrian detector evaluation}
In this section, we highlight the benefit of using LCrowdV to train a learning algorithm
%that extracts features (WHAT FEATURES) from the crowd
and apply the results to pedestrian detection in real videos. 

%\subsubsection{Pedestrian Detection using HOG+SVM}
\noindent{\bf Pedestrian Detection using HOG+SVM:}
We compute the histogram of oriented gradients \cite{1467360} on both positive and negative pedestrian samples in the training dataset as feature descriptor.
%(WHAT ARE THE FEATURE DESCRIPTORS: THE PEDESTRIAN SAMPLES). >>>> I think the HOG feature descriptor looks something like this (right hand side): http://www.mdpi.com/sensors/sensors-13-11603/article_deploy/html/images/sensors-13-11603f3-1024.jpg, not pedestrian samples
We use a support vector machine to learn from these descriptors and learn to determine whether or not a new image patch from the training dataset is a pedestrian or not.  We refer to this method as HOG+SVM.
%IS THIS YOUR NAME OR OTHERS HAVE USED THE SAME METHOD (IF YES, CITE THE OTHER PAPER). >>>>>>>  HOG+SVM is proposed in \cite{1467360}, they mainly talk about HOG, but they used HOG+SVM to show their results in the experiment part, and people later on refer to this method as HOG+SVM/HOG/...
In particular, our SVM detector is trained with OpenCV GPU HOG module and SVM light~\cite{1467360}.
%DO YOU REFERENCES FOR THESE TRAINING METHODS. >>>>>> Yes, \cite{1467360} is the paper describing this method
%\textcolor{red}{[WHEN YOU PERFORM SUCH TRAINING, MENTION THAT YOU CONSIDER OPTIONS OF DATASETS>>>>>>>>Can you elaborate further on this? ]}

\begin{figure}

\vspace{-0.3cm}
\centering
\subfloat[]{\includegraphics[width=0.50\textwidth]{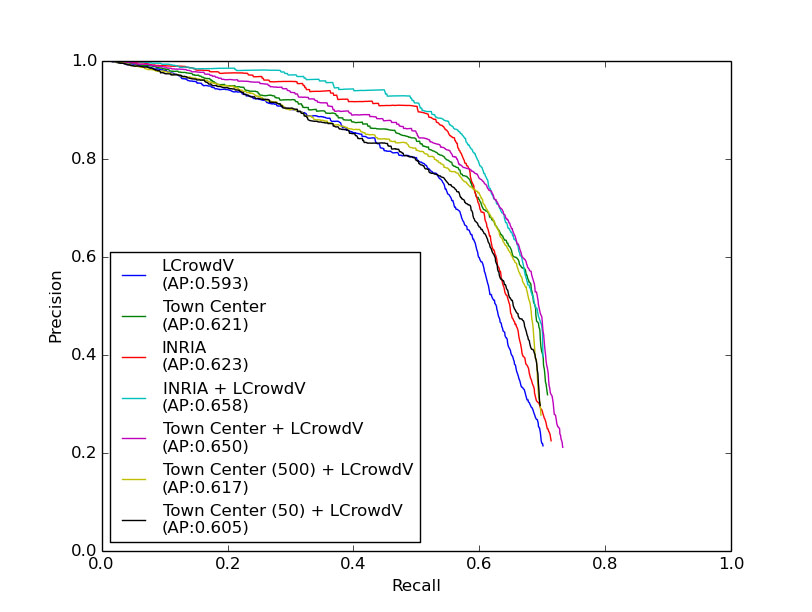}}~
\subfloat[]{\includegraphics[width=0.50\textwidth]{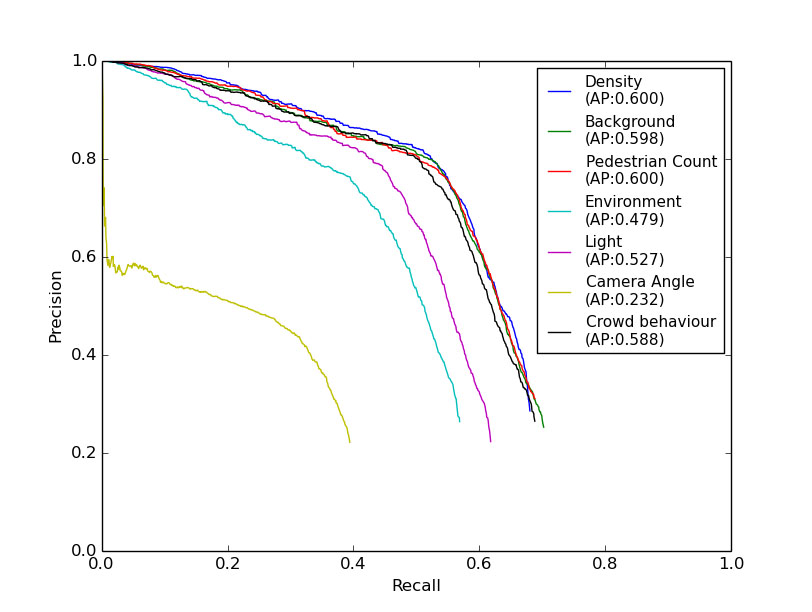}}
\vspace{-1em}
\caption{\small{Results of trained HOG+SVM detectors:  
    (a)  Trained using the realworld and augmented synthetic datasets
    %PUT LABELS FOR AUGMENTED DATASET 
    (INRIA+LCrowdV, Town Center(Full)+LCrowdV, Town Center(500)+LCrowdV and Town Center(50)+LCrowdV)).
%    applied on Town Center Dataset. 
The use of LCrowdV along with the real-world datasets can result in 3\% average precision improvement, as compared to prior results based on only real-world datasets. (b) Different LCrowdV training video datasets obtained by changing each parameter individually. We observe that the variation in the camera angle parameter has maximal impact on improving the accuracy.}}
\label{fig:results}
   \vspace{-0.8cm}
\end{figure}

We trained numerous detectors by combining the real world datasets from INRIA\cite{1467360}, Town Center\cite{benfold2011stable} and our synthetic data in LCrowdV.  
%(ARE THESE IMAGES OR VIDEOS WITH IMAGE SEQUENCES) INRIA are different images, not video
  We used $10,000$ images from our dataset in this experiment.  We observe that the  detectors that are trained by combining LCrowdV and the INRIA or Town Center datasets have higher accuracy as compared to only using the real world datasets (i.e. INRIA only or Town Center only). In  these cases, LCrowdV improves the average precision by ~$3\%$, though we observe higher accuracy for certain cases. %accuracy by $20\%$. 
  We also evaluated two detectors which are trained using only limited samples: $50/500$ positive + $50/500$ negative, from the Town Center datasets and combined with LCrowdV.  The results of the detectors trained by $500+500$ samples are shown to be comparable to the results of the detector trained by the entire original dataset. These benchmarks and results demonstrate that one does not have to spend extensive effort in annotating $70$K image samples to train a detector, merely $1,000$ annotations are sufficient and can be combined with our synthetic LCrowdV dataset. The results are shown in Fig. \ref{fig:results}(a).  In this case, the use of  LCrowdV labeled data can significant improve detectors' accuracy over prior datasets shown in Table \ref{tab:param}.

% \begin{figure}
% \begin{minipage}[b]{0.49\textwidth}
%   \centering
%   \includegraphics[width=\textwidth]{fig8a}
%   \caption{(a)}
%   \label{fig:sfig1}
% \end{subfigure}%
%  \begin{minipage}[b]{0.49\textwidth}
%   \centering
%   \includegraphics[width=\textwidth]{fig8b}
%   \caption{(1)}
%   \label{fig:sfig2}
% \end{subfigure}
% \caption{Results of trained HOG+SVM detectors:  
%     (a)  Trained using the realworld and augmented synthetic datasets
%     %PUT LABELS FOR AUGMENTED DATASET 
%     (INRIA+LCrowdV, Town Center(Full)+LCrowdV, Town Center(500)+LCrowdV and Town Center(50)+LCrowdV)).
% %    applied on Town Center Dataset. 
% The use of LCrowdV along with the real-world datasets can result in 3\% average precision improvement, as compared to prior datasets. (b) Different LCrowdV videos obtained by changing the parameters. We observe that the variation in camera angle has maximal impact on improving the accuracy.}
% \label{fig:results}
% \end{figure}    

\noindent{\bf Varying LCrowdV Parameters:}
%\subsubsection{Varying LCrowdV Parameters} 
Our LCrowdV framework uses seven main parameters,
%(WHERE ARE THESE PARAMETERS DEFINED; PUT A REFERENCE TO THAT SECTION) 
as described in Fig. \ref{fig:tree} and Fig. \ref{fig:label}. We highlight the effect of using different parameters on the accuracy of our detector. 
%on the results, we should ideally test the variations of all combinations of the 7 parameters.  However, to avoid the curse of dimensionality, \textcolor{blue}{we prefer not to test all combinations of variations of all 7 parameters, which leads to computation of $\sum\nolimits_{1}^{7} C(7,i) = 127$ detectors.}
We first train HOG+SVM using a set of synthetic dataset that is generated with variations all seven parameters.  Next, we remove the variations in one parameter at a time and repeat the evaluation.  The results are shown in  Fig. \ref{fig:results}(b).  
%Most numeric parameters are uniformly sampled; The viewpoints are manually or randomly specified using any sampling scheme.

Among the seven parameters, we observe that the variations in the camera angle parameter can affect the average precision by $36\%$, 
%accuracy by $50\%$
 as compared to the other parameters used in LCrowdV. While it is difficult to capture videos from multiple camera angles in real-world scenarios, it is rather simple to vary these parameters in LCrowdV. These results highlight the benefits of LCrowdV.

%upon fixing environment, lighting conditions and camera angle, the accuracy of the trained dataset has been significantly affected. 
%Camera Angle
%\textcolor{blue}{In particular, without the vairation in camera angle, the accruacy of the detector trained dropped by more than 50%.  }
%This has shown the importance of including the variations in the generation of our dataset.  On the other hand, pedestrian count, density, background and crowd behavior has shown minimal effects towards the results.  

%The reason of fixed density and pedestrian count does not have a big impact on the results is that it does not vary much in the testing data.  As for the background, we include real video background as a parameter to provide an option to train scene-specific data.  Yet, we did not extract the background in this test dataset and generate simulated video in this training.  Lastly, for crowd behavior, it is because we are merely using the images to train the pedestrian detectors, and the effect of crowd behavior will become more significant when the data is used as videos.  

%\subsubsection{Pedestrian Detection using Faster R-CNN}
\noindent{\bf Pedestrian Detection using Faster R-CNN:}
%EXPLAIN WHAT IS THE PROBLEM AND WHAT DOES R-CNN REFERS? WHAT IS THE APPLICATION AND HOW CNN's ARE USED? GIVEN SUFFICIENT BACKGROUND
Apart from HOG+SVM, we have also used LCrowdV to train Faster Region-based Convolutional Network method (Faster R-CNN) \cite{renNIPS15fasterrcnn}, one of the state-of-the-art algorithms for object detection based on deep learning. 
R-CNN \cite{girshick2014rich} is a convolutional neural network that makes use of region classification, and it has strong performance in terms of object detection.  A variant, Fast R-CNN\cite{Girshick_2015_ICCV}, combines several ideas to improve the training and testing speed while also increases detection accuracy. 
We use a version of the Fast R-CNN algorithm that makes use of Region Proposal Network (RPN) to improve the performance, namely Faster R-CNN.
% R-CNN and Fast R-CNN and Faster R-CNN are names for 3 different work (all cited)
The RPN makes use of a shared set of convolution layers with the Faster R-CNN network to save computation effort.  In particular, we use the Simonyan and Zisserman model\cite{DBLP:journals/corr/SimonyanZ14a} (VGG-16) that is a very deep detection network and has $13$ shareable convolutional layers.  We adopt the Approximate joint training solution that makes it possible to merge RPN and Fast R-CNN network efficently.  In our implementation, we make use of the Caffe deep learning network\cite{DBLP:journals/corr/JiaSDKLGGD14}, and we iteratively train the model until the performance converges at roughly $10$k to $30$k iterations. 

\begin{figure}
\centering
\vspace{-2em}
\subfloat[]{\includegraphics[width=0.50\textwidth]{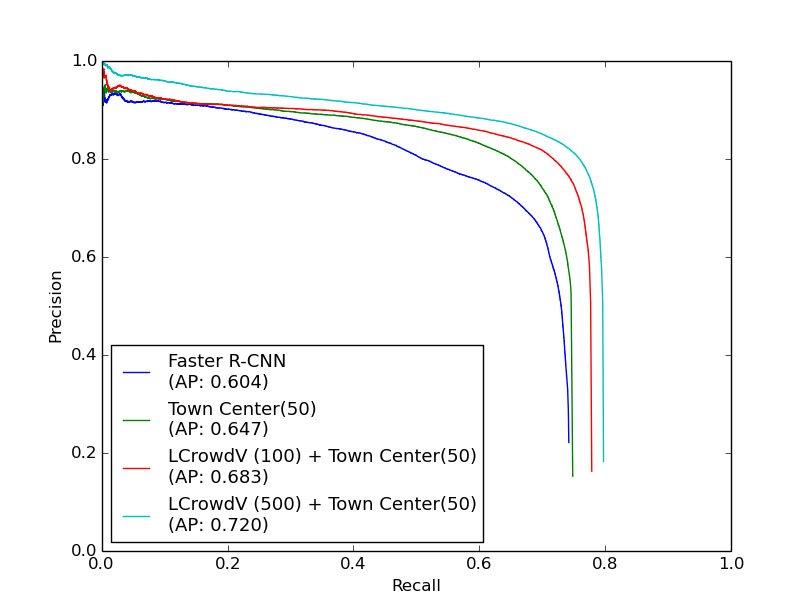}}
\subfloat[]{\includegraphics[width=0.50\textwidth]{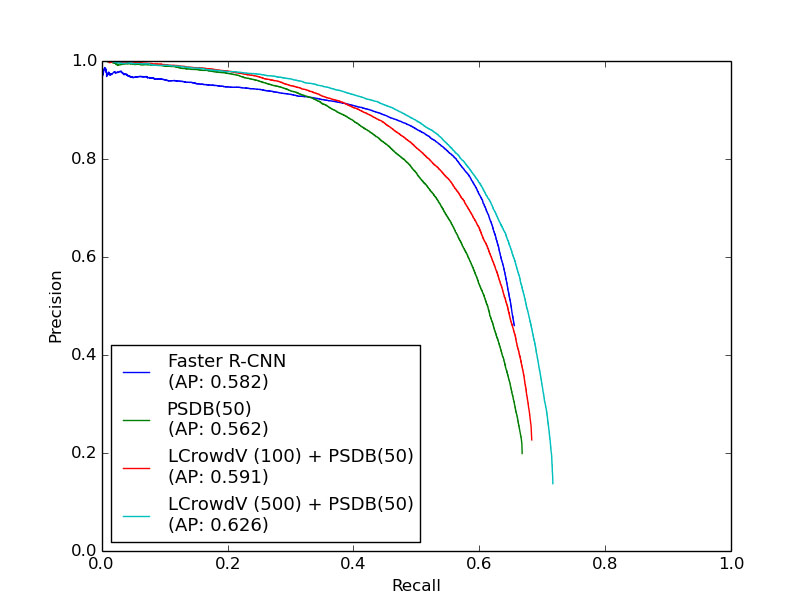}}
\\\vspace{-1.35em}
\subfloat[]{\includegraphics[width=0.50\textwidth]{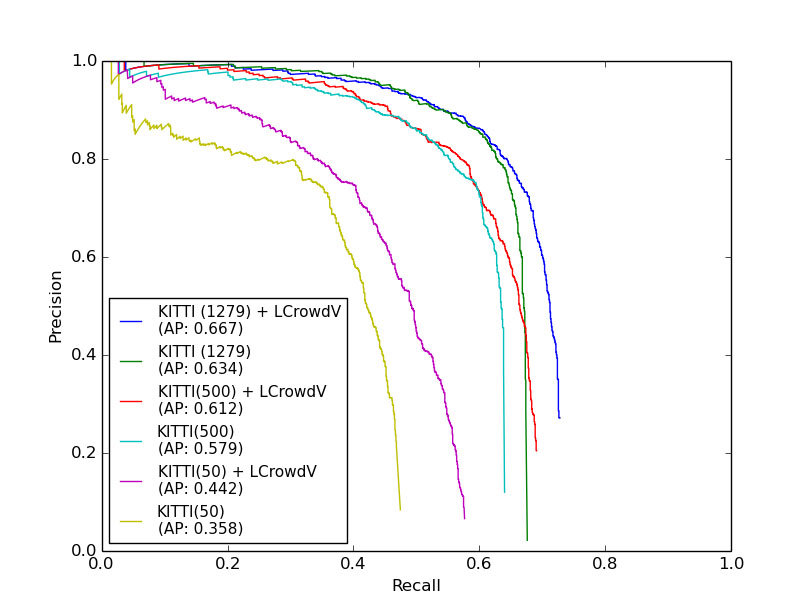}}
\subfloat[]{\includegraphics[width=0.50\textwidth]{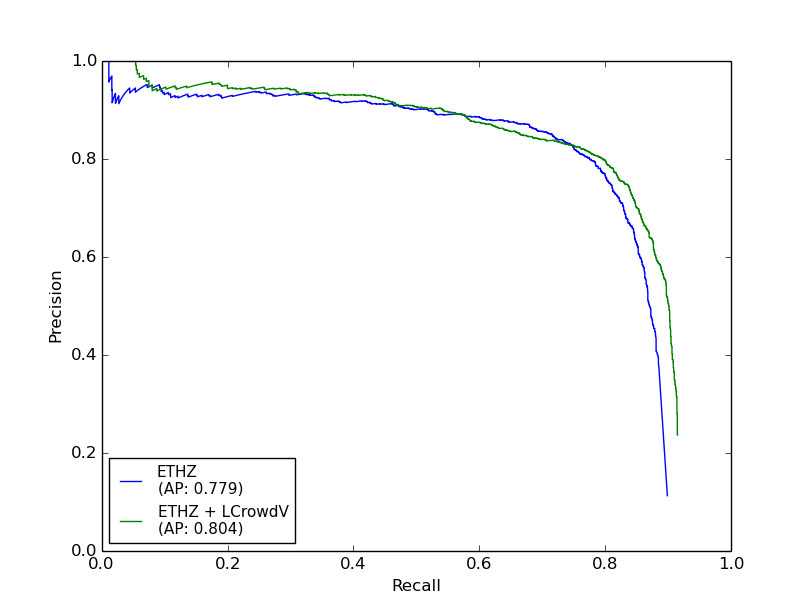}}
\vspace{-1em}
\caption{\small{Results of trained Faster R-CNN model with (a) Town Center dataset, (b) PSDB, (c) KITTI, and (d) ETHZ, and the augmented version of the aforementioned datasets using LCrowdV. The model trained with augmented dataset has an improvement in average precision up to $7.3\%$, $6.4\%$, $3.3\%$ and $2.5\%$ comparing to the model trained with the original dataset for (a), (b), (c), and (d) respectively.  } }
\label{fig:deep}
   \vspace{-0.5cm}
\end{figure}

We trained the model with an augmented dataset which combines both a small sample of Town Center dataset and LCrowdV, and then we use the model to detect pedestrians on the Town Center dataset.  The results are shown in Fig.  \ref{fig:deep}(a).  With merely $50$ samples annotated in the original Town Center dataset, adding LCrowdV into the training set results in an average precision of $72\%$, which is $7.3\%$ better than the model trained with Town Center dataset only.  In addition, we also verify our results by combining LCrowdV with PSDB, KITTI, ETHZ.  For PSDB, the results are shown in Fig. \ref{fig:deep}(b) where the average precision improvement of $6.4\%$ is observed in our combined training set, when comparing to training with samples from PSDB only.  In both experiments, we can observe that as the sample size of LCrowdV increases, the performance of the model becomes better.  

When we evaluate our results on KITTI, we vary the sample size of real annotations to find out also its impact on the performance.  When the number of images with real annotations is \{50, 125, 250, 500, 750, 1000, 1279\}, the AP of KITTI and KITTI+LCrowdV is \{35.8, 48.1, 54.9, 57.9, 61.6, 62.0, 63.4\}\% and \{36.3, 48.9, 55.7, 58.6, 62.7, 64.9, 66.7\}\%, respectively.  The summary of this result is also shown in Fig. \ref{fig:deep}(c).  We can see the complementary effect of LCrowdV on the training is consistently beneficial as the sample size of real annotation varies.  We further evaluate the results on a cross-scene scenario using ETHZ dataset, the improvement of the model trained with combined data is 2.5\% as shown in Fig \ref{fig:deep}(d).  

%This database consists of 18,184 images taken at different angles. And unlike the Town Center dataset which consist of images at the same viewpoint and scene, the scene of PSDB is more diverse, like shopping mall, roadside, University, park, etc.  

The results from both techniques for pedestrian detection mentioned above demonstrate that by combining a small set of samples from the same scene as the test data with LCrowdV, we can improve the detector/deep model results significantly.

% \begin{figure}
%     \centering
%     \begin{minipage}[b]{0.49\textwidth}
%     \centering
%     \includegraphics[width=\textwidth]{fig9}
%     \subcaption{(a)}
%     \end{minipage}
%     \hfill
%     \begin{minipage}[b]{0.49\textwidth}
%     \centering
%      \centering
%     \includegraphics[width=\textwidth]{fig9b}
%     \subcaption{(b)}
%     \end{minipage}

%     \caption{Results of trained Faster R-CNN model with Town Center dataset alone, combined data of Town Center and LCrowdV, and the original Faster R-CNN model.  }
%     \label{fig:deep}
% \end{figure}

\section{Limitations, Conclusion and Future Work}
We have presented a novel approach to generate labeled crowd videos (LCrowdV) using procedural modeling and rendering techniques. Our formulation is general and can be used to include arbitrary numbers of pedestrians, density, behaviors, flows, rendering conditions, and vary the resolution of the images or video. As compared to prior crowd datasets, our synthetic methods can generate a significantly much larger collection of crowd videos with accurate labels or ground truth data. 
We have demonstrated the benefits of LCrowdV in augmenting real world dataset for pedestrian detections. The main benefit of LCrowdV came from easily generated videos from a different camera angle.
 Similarly, we observe considerable improvements in crowd behavior classification because LCrowdV has a large set of labelled videos corresponding to different behavior.
 
Our approach has a few limitations. The current simulation methods may not be able to capture all the details or subtle aspects of human behaviors or movements in certain situations. Our current rendering framework uses the capabilities of Unreal game engine, which may not be able to accurately render many outdoor effects.
In terms of future work, we would like to overcome these limitations. 
%Furthermore, we would like to evaluate the benefits in terms of improved algorithms for crowd counting, tracking, abnormal behavior detection, and segmentation. 
  We would like to continue investigating in improving machine learning algorithms in realted to crowds, including crowd counting, tracking, abnormal behavior detection, crowd segmentation and etc.
We would also like to include traffic in these videos and generate datasets corresponding to human-vehicle interactions. We also plan to make the LCrowdV dataset available on the WWW.
%\textcolor{blue}{Should this work be published, we will also release the dataset on Internet, and we welcome the community to join us in the aforementioned studies on further applications of LCrowdV.}

\bibliographystyle{splncs}
\bibliography{egbib}

\begin{thebibliography}{10}

\bibitem{mikel_dataDriven}
Rodriguez, M., Sivic, J., Laptev, I., Audibert, J.Y.:
\newblock Data-driven crowd analysis in videos.
\newblock In: 2011 International Conference on Computer Vision. (Nov 2011)
  1235--1242

\bibitem{rabaud2006counting}
Rabaud, V., Belongie, S.:
\newblock Counting crowded moving objects.
\newblock In: Computer Vision and Pattern Recognition, 2006 IEEE Computer
  Society Conference on. Volume~1., IEEE (2006)  705--711

\bibitem{hattori2015learning}
Hattori, H., Boddeti, V.N., Kitani, K., Kanade, T.:
\newblock Learning scene-specific pedestrian detectors without real data.
\newblock In: Proceedings of the IEEE Conference on Computer Vision and Pattern
  Recognition. (2015)  3819--3827

\bibitem{5597451}
Ozturk, O., Yamasaki, T., Aizawa, K.:
\newblock Detecting dominant motion flows in unstructured/structured crowd
  scenes.
\newblock In: Pattern Recognition (ICPR), 2010 20th International Conference
  on. (Aug 2010)  3533--3536

\bibitem{shao2015deeply}
Shao, J., Kang, K., Loy, C.C., Wang, X.:
\newblock Deeply learned attributes for crowded scene understanding.
\newblock In: 2015 IEEE Conference on Computer Vision and Pattern Recognition
  (CVPR). (June 2015)  4657--4666

\bibitem{AliFlow}
Ali, S., Shah, M.:
\newblock A lagrangian particle dynamics approach for crowd flow segmentation
  and stability analysis.
\newblock In: Computer Vision and Pattern Recognition, 2007. CVPR 2007. IEEE
  Conference on. (June 2007)  1--6

\bibitem{marin2010learning}
Marín, J., Vázquez, D., Gerónimo, D., López, A.M.:
\newblock Learning appearance in virtual scenarios for pedestrian detection.
\newblock In: Computer Vision and Pattern Recognition (CVPR), 2010 IEEE
  Conference on. (June 2010)  137--144

\bibitem{dhome1993determination}
Dhome, M., Yassine, A., Lavest, J.M.:
\newblock Determination of the pose of an articulated object from a single
  perspective view.
\newblock In: BMVC. (1993)  1--10

\bibitem{movshovitz20143d}
Movshovitz-Attias, Y., Boddeti, V.N., Wei, Z., Sheikh, Y.:
\newblock 3d pose-by-detection of vehicles via discriminatively reduced
  ensembles of correlation filters.
\newblock In: British Machine Vision Conference. (2014)

\bibitem{pepik2012teaching}
Pepik, B., Stark, M., Gehler, P., Schiele, B.:
\newblock Teaching 3d geometry to deformable part models.
\newblock In: Computer Vision and Pattern Recognition (CVPR), 2012 IEEE
  Conference on. (June 2012)  3362--3369

\bibitem{satkin2012data}
Satkin, S., Lin, J., Hebert, M.:
\newblock Data-driven scene understanding from 3d models.
\newblock (2012)

\bibitem{ali2013modeling}
Ali, S., Nishino, K., Manocha, D., Shah, M.:
\newblock Modeling, Simulation and Visual Analysis of Crowds: A
  Multidisciplinary Perspective.
\newblock Springer (2013)

\bibitem{zhou2012understanding}
Zhou, B., Wang, X., Tang, X.:
\newblock Understanding collective crowd behaviors: Learning a mixture model of
  dynamic pedestrian-agents.
\newblock In: Computer Vision and Pattern Recognition (CVPR), 2012 IEEE
  Conference on. (June 2012)  2871--2878

\bibitem{hospedales2009markov}
Hospedales, T., Gong, S., Xiang, T.:
\newblock A markov clustering topic model for mining behaviour in video.
\newblock In: Computer Vision, 2009 IEEE 12th International Conference on, IEEE
  (2009)  1165--1172

\bibitem{Mehran_NormalAbnormalSocialForce}
Mehran, R., Oyama, A., Shah, M.:
\newblock Abnormal crowd behavior detection using social force model.
\newblock In: Computer Vision and Pattern Recognition, 2009. CVPR 2009. IEEE
  Conference on. (June 2009)  935--942

\bibitem{4359353}
Chan, A., Vasconcelos, N.:
\newblock Modeling, clustering, and segmenting video with mixtures of dynamic
  textures.
\newblock Pattern Analysis and Machine Intelligence, IEEE Transactions on
  \textbf{30}(5) (May 2008)  909--926

\bibitem{4731265}
Wang, X., Ma, X., Grimson, W.:
\newblock Unsupervised activity perception in crowded and complicated scenes
  using hierarchical bayesian models.
\newblock Pattern Analysis and Machine Intelligence, IEEE Transactions on
  \textbf{31}(3) (March 2009)  539--555

\bibitem{bera2016realtime}
Bera, A., Kim, S., Manocha, D.:
\newblock Realtime anomaly detection using trajectory-level crowd behavior
  learning.
\newblock In: Proceedings of the IEEE Conference on Computer Vision and Pattern
  Recognition Workshops. (2016)  50--57

\bibitem{kim2015interactive}
Kim, S., Bera, A., Manocha, D.:
\newblock Interactive crowd content generation and analysis using
  trajectory-level behavior learning.
\newblock In: 2015 IEEE International Symposium on Multimedia (ISM), IEEE
  (2015)  21--26

\bibitem{5540833}
Xu, H., Lv, P., Meng, L.:
\newblock A people counting system based on head-shoulder detection and
  tracking in surveillance video.
\newblock In: Computer Design and Applications (ICCDA), 2010 International
  Conference on. Volume~1. (June 2010)  V1--394--V1--398

\bibitem{1683826}
Antonini, G., Thiran, J.P.:
\newblock Counting pedestrians in video sequences using trajectory clustering.
\newblock Circuits and Systems for Video Technology, IEEE Transactions on
  \textbf{16}(8) (Aug 2006)  1008--1020

\bibitem{garcia2009density}
Garcia-Bunster, G., Torres-Torriti, M.:
\newblock A density-based approach for effective pedestrian counting at bus
  stops.
\newblock In: Systems, Man and Cybernetics, 2009. SMC 2009. IEEE International
  Conference on, IEEE (2009)  3434--3439

\bibitem{idrees2013multi}
Idrees, H., Saleemi, I., Seibert, C., Shah, M.:
\newblock Multi-source multi-scale counting in extremely dense crowd images.
\newblock In: Computer Vision and Pattern Recognition (CVPR), 2013 IEEE
  Conference on. (June 2013)  2547--2554

\bibitem{zhang2015cross}
Zhang, C., Li, H., Wang, X., Yang, X.:
\newblock Cross-scene crowd counting via deep convolutional neural networks.
\newblock In: 2015 IEEE Conference on Computer Vision and Pattern Recognition
  (CVPR). (June 2015)  833--841

\bibitem{zhang2012real}
Zhang, K., Zhang, L., Yang, M.H.:
\newblock Real-time compressive tracking.
\newblock In: ECCV 2012.
\newblock (2012)  864--877

\bibitem{Fu2012}
Fu, W., Wang, J., Li, Z., Lu, H., Ma, S.:
\newblock Learning semantic motion patterns for dynamic scenes by improved
  sparse topical coding.
\newblock In: Multimedia and Expo (ICME), IEEE International Conference on.
  (2012)  296--301

\bibitem{song2013fully}
Song, X., Shao, X., Zhang, Q., Shibasaki, R., Zhao, H., Cui, J., Zha, H.:
\newblock A fully online and unsupervised system for large and high-density
  area surveillance: Tracking, semantic scene learning and abnormality
  detection.
\newblock TIST (2013)

\bibitem{ali2008floor}
Ali, S., Shah, M.:
\newblock Floor fields for tracking in high density crowd scenes.
\newblock In: ECCV 2008.
\newblock Springer (2008)  1--14

\bibitem{zhu2014crowd}
Zhu, F., Wang, X., Yu, N.:
\newblock Crowd tracking with dynamic evolution of group structures.
\newblock In: ECCV 2014.
\newblock Springer (2014)  139--154

\bibitem{LiCrowdedSceneAnalysis2015}
Li, T., Chang, H., Wang, M., Ni, B., Hong, R., Yan, S.:
\newblock Crowded scene analysis: A survey.
\newblock Circuits and Systems for Video Technology, IEEE Transactions on
  \textbf{25}(3) (March 2015)  367--386

\bibitem{solmaz2012identifying}
Solmaz, B., Moore, B.E., Shah, M.:
\newblock Identifying behaviors in crowd scenes using stability analysis for
  dynamical systems.
\newblock Pattern Analysis and Machine Intelligence, IEEE Transactions on
  \textbf{34}(10) (2012)  2064--2070

\bibitem{kratz2009anomaly}
Kratz, L., Nishino, K.:
\newblock Anomaly detection in extremely crowded scenes using spatio-temporal
  motion pattern models.
\newblock In: Computer Vision and Pattern Recognition. IEEE Conference on, IEEE
  (2009)  1446--1453

\bibitem{6027295}
Srivastava, S., Ng, K., Delp, E.:
\newblock Crowd flow estimation using multiple visual features for scenes with
  changing crowd densities.
\newblock In: Advanced Video and Signal-Based Surveillance (AVSS), 2011 8th
  IEEE International Conference on. (Aug 2011)  60--65

\bibitem{chen2006intelligent}
Chen, T.H., Chen, T.Y., Chen, Z.X.:
\newblock An intelligent people-flow counting method for passing through a
  gate.
\newblock In: Robotics, Automation and Mechatronics, 2006 IEEE Conference on,
  IEEE (2006)  1--6

\bibitem{tsuduki2009method}
Tsuduki, Y., Fujiyoshi, H.:
\newblock A method for visualizing pedestrian traffic flow using sift feature
  point tracking.
\newblock In: Advances in Image and Video Technology.
\newblock Springer (2009)  25--36

\bibitem{5206827}
Loy, C.C., Xiang, T., Gong, S.:
\newblock Multi-camera activity correlation analysis.
\newblock In: Computer Vision and Pattern Recognition, 2009. CVPR 2009. IEEE
  Conference on. (June 2009)  1988--1995

\bibitem{1698861}
Andrade, E., Blunsden, S., Fisher, R.:
\newblock Modelling crowd scenes for event detection.
\newblock In: Pattern Recognition, 2006. ICPR 2006. 18th International
  Conference on. Volume~1. (2006)  175--178

\bibitem{5653573}
Dee, H., Caplier, A.:
\newblock Crowd behaviour analysis using histograms of motion direction.
\newblock In: Image Processing (ICIP), 2010 17th IEEE International Conference
  on. (Sept 2010)  1545--1548

\bibitem{rodriguez2008action}
Rodriguez, M.D., Ahmed, J., Shah, M.:
\newblock Action mach a spatio-temporal maximum average correlation height
  filter for action recognition.
\newblock In: Computer Vision and Pattern Recognition, 2008. CVPR 2008. IEEE
  Conference on. (June 2008)  1--8

\bibitem{reddy2013recognizing}
Reddy, K.K., Shah, M.:
\newblock Recognizing 50 human action categories of web videos.
\newblock Machine Vision and Applications \textbf{24}(5) (2013)  971--981

\bibitem{DBLP:journals/corr/abs-1212-0402}
Soomro, K., Zamir, A.R., Shah, M.:
\newblock {UCF101:} {A} dataset of 101 human actions classes from videos in the
  wild.
\newblock CoRR \textbf{abs/1212.0402} (2012)

\bibitem{6909682}
Shao, J., Loy, C., Wang, X.:
\newblock Scene-independent group profiling in crowd.
\newblock In: Computer Vision and Pattern Recognition (CVPR), 2014 IEEE
  Conference on. (June 2014)  2227--2234

\bibitem{6714561}
Zhou, B., Tang, X., Zhang, H., Wang, X.:
\newblock Measuring crowd collectiveness.
\newblock Pattern Analysis and Machine Intelligence, IEEE Transactions on
  \textbf{36}(8) (Aug 2014)  1586--1599

\bibitem{6239348}
Hassner, T., Itcher, Y., Kliper-Gross, O.:
\newblock Violent flows: Real-time detection of violent crowd behavior.
\newblock In: Computer Vision and Pattern Recognition Workshops (CVPRW), 2012
  IEEE Computer Society Conference on. (June 2012)  1--6

\bibitem{xu:2014}
J.~Xu, D.~Vazquez, A.M.L.J.M.D.P.:
\newblock Learning a part-based pedestrian detector in virtual world.
\newblock IEEE Transactions on Intelligent Transportation Systems (2014)

\bibitem{Lebon1895}
Le~Bon, G.:
\newblock The crowd: A study of the popular mind.
\newblock Macmillian (1897)

\bibitem{james1953distribution}
James, J.:
\newblock The distribution of free-forming small group size.
\newblock American Sociological Review (1953)

\bibitem{turner1987collective}
Turner, R.H., Killian, L.M.:
\newblock Collective behavior (pp. 1--14, 16) (1987)

\bibitem{pervin2003science}
Pervin, L.:
\newblock The Science of Personality.
\newblock Oxford University Press (2003)

\bibitem{moussaid2010walking}
Moussa{\"\i}d, M., Perozo, N., Garnier, S., Helbing, D., Theraulaz, G.:
\newblock The walking behaviour of pedestrian social groups and its impact on
  crowd dynamics.
\newblock (2010)

\bibitem{eysenck1985personality}
Eysenck, H., Eysenck, M.:
\newblock Personality and individual differences: A natural science perspective
  (1985)

\bibitem{curtis2014menge}
Curtis, S., Best, A., Manocha, D.:
\newblock Menge: A modular framework for simulating crowd movement.
\newblock Technical report Department of Computer Science, University of North
  Carolina at Chapel Hill.

\bibitem{funge99}
Funge, J., Tu, X., Terzopoulos, D.:
\newblock Cognitive modeling: knowledge, reasoning and planning for intelligent
  characters.
\newblock In: Proceedings of the 26th annual conference on Computer graphics
  and interactive techniques, ACM Press/Addison-Wesley Publishing Co. (1999)
  29--38

\bibitem{ulicny2002towards}
Ulicny, B., Thalmann, D.:
\newblock Towards interactive real-time crowd behavior simulation.
\newblock In: Computer Graphics Forum. Volume~21., Wiley Online Library (2002)
  767--775

\bibitem{shao2005autonomous}
Shao, W., Terzopoulos, D.:
\newblock Autonomous pedestrians.
\newblock In: Proceedings of the 2005 ACM SIGGRAPH/Eurographics symposium on
  Computer animation, ACM (2005)  19--28

\bibitem{barraquand1991robot}
Barraquand, J., Latombe, J.C.:
\newblock Robot motion planning: A distributed representation approach.
\newblock The International Journal of Robotics Research \textbf{10}(6) (1991)
  628--649

\bibitem{snook00}
Snook, G.:
\newblock Simplified 3d movement and pathfinding using navigation meshes.
\newblock In DeLoura, M., ed.: Game Programming Gems.
\newblock Charles River Media (2000)  288--304

\bibitem{lamarche2004crowd}
Lamarche, F., Donikian, S.:
\newblock Crowd of virtual humans: a new approach for real time navigation in
  complex and structured environments.
\newblock In: Computer Graphics Forum. Volume~23., Wiley Online Library (2004)
  509--518

\bibitem{geraerts2008using}
Geraerts, R., Kamphuis, A., Karamouzas, I., Overmars, M.:
\newblock Using the corridor map method for path planning for a large number of
  characters.
\newblock In: Motion in Games.
\newblock Springer (2008)  11--22

\bibitem{boids}
Reynolds, C.:
\newblock Flocks, herds and schools: A distributed behavioral model.
\newblock In: Proc. of SIGGRAPH. (1987)

\bibitem{helbing1995social}
Helbing, D., Molnar, P.:
\newblock Social force model for pedestrian dynamics.
\newblock Physical review E (1995)

\bibitem{Jur:2011:RVO}
van~den Berg, J., Guy, S., Lin, M., Manocha, D.:
\newblock Reciprocal n-body collision avoidance.
\newblock In Pradalier, C., Siegwart, R., Hirzinger, G., eds.: Robotics
  Research. Volume~70 of Springer Tracts in Advanced Robotics.
\newblock Springer Berlin Heidelberg (2011)  3--19

\bibitem{brude89}
Bruderlin, A., Calvert, T.W.:
\newblock Goal-directed, dynamic animation of human walking.
\newblock In: Proc. of SIGGRAPH '89. (1989)  233--242

\bibitem{Lee07}
Lee, K.H., Choi, M.G., Hong, Q., Lee, J.:
\newblock Group behavior from video: a data-driven approach to crowd
  simulation.
\newblock In: Symposium on Computer Animation. (2007)  109--118

\bibitem{vanB11}
van Basten, B.J.H., Stuvel, S.A., Egges, A.:
\newblock A hybrid interpolation scheme for footprint-driven walking synthesis.
\newblock Graphics Interface (2011)  9--16

\bibitem{Oliver:2012:UEE:2341836.2341909}
Oliver, P.:
\newblock Unreal engine 4 elemental.
\newblock In: ACM SIGGRAPH 2012 Computer Animation Festival. SIGGRAPH '12, New
  York, NY, USA, ACM (2012)  86--86

\bibitem{1467360}
Dalal, N., Triggs, B.:
\newblock Histograms of oriented gradients for human detection.
\newblock In: Computer Vision and Pattern Recognition, 2005. CVPR 2005. IEEE
  Computer Society Conference on. Volume~1. (June 2005)  886--893 vol. 1

\bibitem{6248074}
Geiger, A., Lenz, P., Urtasun, R.:
\newblock Are we ready for autonomous driving? the kitti vision benchmark
  suite.
\newblock In: Computer Vision and Pattern Recognition (CVPR), 2012 IEEE
  Conference on. (June 2012)  3354--3361

\bibitem{4587581}
Ess, A., Leibe, B., Schindler, K., Gool, L.V.:
\newblock A mobile vision system for robust multi-person tracking.
\newblock In: Computer Vision and Pattern Recognition, 2008. CVPR 2008. IEEE
  Conference on. (June 2008)  1--8

\bibitem{benfold2011stable}
Benfold, B., Reid, I.:
\newblock Stable multi-target tracking in real-time surveillance video.
\newblock In: Computer Vision and Pattern Recognition (CVPR), 2011 IEEE
  Conference on. (June 2011)  3457--3464

\bibitem{renNIPS15fasterrcnn}
Ren, S., He, K., Girshick, R., Sun, J.:
\newblock Faster {R-CNN}: Towards real-time object detection with region
  proposal networks.
\newblock In: Advances in Neural Information Processing Systems ({NIPS}).
  (2015)

\bibitem{girshick2014rich}
Girshick, R., Donahue, J., Darrell, T., Malik, J.:
\newblock Rich feature hierarchies for accurate object detection and semantic
  segmentation.
\newblock In: Proceedings of the IEEE conference on computer vision and pattern
  recognition. (2014)  580--587

\bibitem{Girshick_2015_ICCV}
Girshick, R.:
\newblock Fast r-cnn.
\newblock In: 2015 IEEE International Conference on Computer Vision (ICCV).
  (Dec 2015)  1440--1448

\bibitem{DBLP:journals/corr/SimonyanZ14a}
Simonyan, K., Zisserman, A.:
\newblock Very deep convolutional networks for large-scale image recognition.
\newblock CoRR \textbf{abs/1409.1556} (2014)

\bibitem{DBLP:journals/corr/JiaSDKLGGD14}
Jia, Y., Shelhamer, E., Donahue, J., Karayev, S., Long, J., Girshick, R.B.,
  Guadarrama, S., Darrell, T.:
\newblock Caffe: Convolutional architecture for fast feature embedding.
\newblock CoRR \textbf{abs/1408.5093} (2014)

\end{thebibliography}
\end{document}

% \linenumbers
\pagestyle{headings}
\mainmatter
\def\ECCV16SubNumber{497}  % Insert your submission number here

\title{Appendix} % Replace with your title

% \titlerunning{ECCV-16 submission ID \ECCV16SubNumber}

% \authorrunning{ECCV-16 submission ID \ECCV16SubNumber}

\author{Anonymous}
\institute{Paper ID \ECCV16SubNumber}

\maketitle

\section{Crowd Behavior Mapping}

\[
 \begin{pmatrix}
 Aggressive\\
 Assertive\\
 Shy\\
 Active\\
 Tense\\
 Impulsive\\
  \end{pmatrix}
  =A_{adj}
  \begin{pmatrix}
 \frac{1}{13.5}(Neighbor~Dist - 15)\\[0.2em]
 \frac{1}{49.5}(Max.~Neighbors - 10)\\[0.2em]
 \frac{1}{14.5}(Planning~Horiz. - 30)\\[0.2em]
 \frac{1}{0.85}(Radius - 0.8)\\[0.2em]
\frac{1}{0.5}( Pref.~Speed - 1.4)\\[0.2em]
  \end{pmatrix}
  \]
 Matrix $A_{adj}$ is derived using a linear least-squares approach on the user study data we found the following 6-by-5 matrix $A_{adj}$:
\[
 A_{adj} =
 \begin{pmatrix}
  -0.02 & 0.32 & 0.13 & -0.41&1.02\\
  0.03 & 0.22 & 0.11 & -0.28 & 1.05\\
  -0.04 & -0.08 & 0.02 & 0.58 & -0.88\\
  -0.06 & 0.04 & 0.04 & -0.16 & 1.07\\
  0.10 & 0.07 & -0.08 & 0.19 & 0.15\\
  0.03 & -0.15 & 0.03 & -0.23 & 0.23\\
 
 \end{pmatrix}
\]